%% file: main.tex
\PassOptionsToPackage{table}{xcolor}
\documentclass[runningheads]{llncs}

\usepackage[preprint]{colm} 

\setcitestyle{authoryear}

\usepackage{graphicx}
\usepackage{booktabs}
\usepackage{adjustbox}
\usepackage{threeparttable}
\usepackage[accsupp]{axessibility}
\usepackage{xltabular} 
\usepackage{microtype}
\usepackage{tabularx}
\usepackage{amssymb}
\usepackage{paralist}
\usepackage{todonotes}
\usepackage{breakurl}
\usepackage{subcaption}
\usepackage{latexsym}
\usepackage{amsmath}
\usepackage{float}
\usepackage{footnote}
\usepackage{enumitem}
\usepackage{bm}
\usepackage{arydshln}
\usepackage{multicol}
\usepackage{multirow}
\usepackage{colortbl}
\usepackage{bbding}
\usepackage{makecell}
\usepackage{mathtools}
\usepackage{imakeidx}
\usepackage{longtable}
\usepackage{tabularx}
\usepackage{wrapfig}
\usepackage{rotating}
\usepackage[edges]{forest}
\usepackage[normalem]{ulem}
\usepackage{CJKutf8}
\usepackage{awesomebox}
\usepackage{diagbox}
\usepackage[most]{tcolorbox}

\RequirePackage{xspace}
\makeatletter
\DeclareRobustCommand\onedot{\futurelet\@let@token\@onedot}
\def\@onedot{\ifx\@let@token.\else.\null\fi\xspace}

\makeatother

\usepackage{tocloft}

\cftsetindents{section}{0em}{1.5em}
\cftsetindents{subsection}{1.5em}{2.3em}
\cftsetindents{subsubsection}{3.8em}{3.2em}


\usepackage{pifont}
\usepackage{lscape}
\usepackage{algorithm}
\usepackage{algpseudocode}

\usepackage{fancyhdr}
\renewcommand{\headrulewidth}{1pt}
\usepackage{fancyvrb}
\usepackage{fvextra}
\usepackage{amsfonts}
\usepackage{wrapfig}
\usepackage{setspace}
\usepackage{hyperref}
\usepackage{orcidlink}

\definecolor{mydarkblue}{rgb}{0,0.08,0.45}
\definecolor{wkblue}{rgb}{0.2, 0.3, 0.6}
\definecolor{meta-color}{rgb}{0.5, 0.5, 0.5}
\definecolor{darkblue}{rgb}{0, 0, 0.5}
\definecolor{geovistagray}{gray}{0.95}
\definecolor{myblue}{rgb}{0.9, 0.1, 0.94}
\definecolor{mygreen}{rgb}{0.64, 0.56, 0.88}
\definecolor{myyellow}{rgb}{0.68, 0.6, 0.1}
\definecolor{fancygreen}{rgb}{0.33, 0.68, 0.20}
\definecolor{salmon}{rgb}{0.94, 0.52, 0.49}
\definecolor{tablegreen}{rgb}{0.82, 0.94, 0.75}
\definecolor{tableblue}{rgb}{0.81, 0.90, 0.94}
\definecolor{tablered}{rgb}{0.97, 0.85, 0.85}
\definecolor{tableorange}{rgb}{0.96, 0.85, 0.81}
\definecolor{bestcolor}{RGB}{210, 222, 239}
\definecolor{secondcolor}{RGB}{234, 239, 247}
\definecolor{thirdcolor}{RGB}{193, 214, 229}
\definecolor{line-blue}{RGB}{243, 248, 252}
\definecolor{line-green}{RGB}{200,242,200}
\definecolor{line-red}{RGB}{255,215,215}
\definecolor{line-gray}{RGB}{242, 242, 242}
\definecolor{sensepurple}{HTML}{5D2DD6}
\definecolor{rynn}{RGB}{108,92,186} 

\newenvironment{itemize*}%
 {\leftmargini=10pt\begin{itemize}%
  \setlength{\itemsep}{0pt}%
  \setlength{\parskip}{0pt}%
  }%
 {\end{itemize}}
\newenvironment{enumerate*}%
 {\begin{enumerate}%
  \setlength{\itemsep}{0pt}%
  \setlength{\parskip}{0pt}}%
 {\end{enumerate}}

\hypersetup{
    hidelinks
}

\newcommand{\paragrapha}[2][3pt]{\vspace{#1}\noindent\textbf{#2}}

\newcolumntype{x}[1]{>{\centering\arraybackslashå}p{#1pt}}
\newlength\savewidth

\newcommand{\PreserveBackslash}[1]{\let\temp=\\#1\let\\=\temp}
\newcolumntype{C}[1]{>{\PreserveBackslash\centering}p{#1}}
\newcolumntype{L}[1]{>{\PreserveBackslash\raggedright}p{#1}}

\usepackage[dvipsnames]{xcolor} 
\usepackage{hyperref}
\hypersetup{
    colorlinks=true,
    linkcolor=RoyalBlue,
    citecolor=RoyalBlue,
    urlcolor=MidnightBlue
}

\begin{document}

\title{HY-Embodied-0.5: Embodied Foundation Models for \\Real-World Agents}

\titlerunning{HY-Embodied}

\author{\textbf{Tencent Robotics X} ${\times}$ \textbf{HY Vision Team}}

\authorrunning{}

\maketitle

\makeatletter
\def\@makefnmark{}
\makeatother


\input{sec/0-Abstract}

\newpage
\tableofcontents
\newpage

\input{sec/1-Introduction}
\input{sec/2-Architecture}
\input{sec/3-Pretrain}
\input{sec/4-Posttrain1}
\input{sec/5-Posttrain2}
\input{sec/6-Evaluation}

\input{sec/7-VLA}
\input{sec/8-Conclusion}

\newpage

\renewcommand{\refname}{References}
\renewcommand{\bibname}{References}
\renewcommand{\bibsection}{\section*{\raggedright \Large References}}
\makeatother
\bibliographystyle{abbrvnat}
\bibliography{egbib}

\newpage

\appendix
\section{Contributors}

\vspace{1em}
\begin{itemize}[label=$\bullet$, leftmargin=*, itemsep=0.8em]
\item \textbf{Project Sponsors:} Zhengyou Zhang, Linus, Shunyu Yao
\item \textbf{Project Supervisor:} Han Hu
\item \textbf{Project Leader:} Yongming Rao
\item \textbf{Core Contributors:} Xumin Yu, Zuyan Liu, Ziyi Wang, He Zhang
\item \textbf{Contributors:} Fangfu Liu, Yani Zhang, Ruowen Zhao, Oran Wang, Yves Liang, Haitao Lin, Minghui  Wang, Yubo Dong, Kevin Cheng, Bolin Ni, Rui Huang
\end{itemize}

\newpage
\input{sec/9-Appendix}
\end{document}

%% file: sec/0-Abstract.tex
\begin{abstract}
We introduce HY-Embodied-0.5, a family of foundation models specifically designed for real-world embodied agents. To bridge the gap between general Vision-Language Models (VLMs) and the demands of embodied agents, our models are developed to enhance the core capabilities required by embodied intelligence: spatial and temporal visual perception, alongside advanced embodied reasoning for prediction, interaction, and planning. The HY-Embodied-0.5 suite comprises two primary variants: an efficient model with 2B activated parameters designed for edge deployment, and a powerful model with 32B activated parameters targeted for complex reasoning. To support the fine-grained visual perception essential for embodied tasks, we adopt a Mixture-of-Transformers (MoT) architecture to enable modality-specific computing. By incorporating latent tokens, this design effectively enhances the perceptual representation of the models. To improve reasoning capabilities, we introduce an iterative, self-evolving post-training paradigm. Furthermore, we employ on-policy distillation to transfer the advanced capabilities of the large model to the smaller variant, thereby maximizing the performance potential of the compact model. Extensive evaluations across 22 benchmarks, spanning visual perception, spatial reasoning, and embodied understanding, demonstrate the effectiveness of our approach. Our MoT-2B model outperforms similarly sized state-of-the-art models on 16 benchmarks, while the 32B variant achieves performance comparable to frontier models such as Gemini 3.0 Pro. In downstream robot control experiments, we leverage our robust VLM foundation to train an effective Vision-Language-Action (VLA) model, achieving compelling results in real-world physical evaluations. Code and models are open-sourced at \url{https://github.com/Tencent-Hunyuan/HY-Embodied}.
\end{abstract}

\begin{figure}[htbp]
  \centering
  \includegraphics[width=\linewidth]{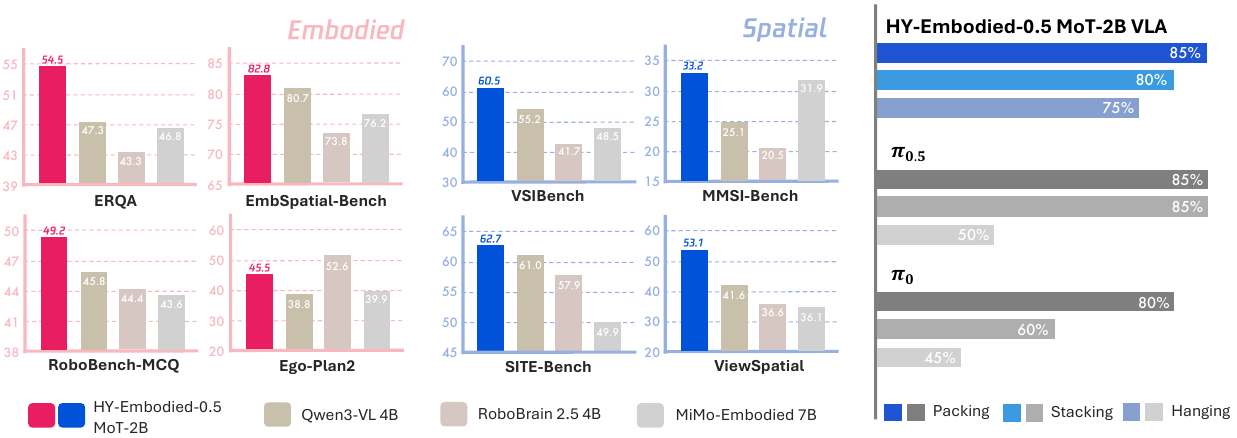}
  \caption{\textbf{Performance of HY-Embodied-0.5 MoT-2B on spatial and embodied benchmarks as well as downstream robot control tasks. } HY-Embodied-0.5 pushes the frontier of embodied VLMs, while excelling in downstream real-world robot evaluations. }
  \label{fig:1-teaser}
\end{figure}

%% file: sec/1-Introduction.tex
\section{Introduction}

Intelligent agents~\citep{yao2023react,park2023generative,yang2024sweagent,xie2024osworld} have emerged as a foundational paradigm for problem-solving~\citep{lu2024aiscientist} and workflow automation~\citep{yang2024sweagent,wu2024oscopilot}, driven by revolutionary progress in large language models (LLMs)~\citep{gemini3,openai2025gpt5,anthropic2025claude4,deepseek2024v3}. These agents are playing an increasingly pivotal role in diverse applications, ranging from coding~\citep{yang2024sweagent} and personal assistance~\citep{wu2024oscopilot} to scientific research~\citep{lu2024aiscientist}. Extending agents into physical environments naturally becomes a promising next frontier. While Vision-Language Models (VLMs)~\citep{liu2024llava,bai2023qwenvl} have achieved substantive progress in recent years, developing VLMs capable of seamlessly perceiving, reasoning, and acting within physical and embodied scenarios remains a significant challenge. To enable robust real-world agents, current VLMs require substantial advancements in two primary dimensions: \textbf{(1) Fine-Grained Visual Perception:} Precise, fine-grained visual perception is the fundamental prerequisite for understanding the physical world and making informed decisions for specific actions. However, existing VLMs still exhibit notable deficiencies in capturing the granular details required for physical grounding. \textbf{(2) Embodied Prediction, Interaction, and Planning:} Mainstream VLMs, predominantly trained on static, web-scale datasets, excel in general-purpose scenarios but remain inadequately optimized for embodied environments, lacking the action-oriented capabilities essential for dynamic prediction, interaction, and planning in the physical world.

In this report, we present HY-Embodied-0.5, a family of foundation models purpose-built for real-world agents. Driven by the goal of translating digital intelligence into the physical world, we build these models upon the VLM paradigm~\citep{liu2024llava}. We believe that embodied VLMs uniquely bridge the gap between LLM agents and physical agents, enabling the vast open-world knowledge of large language and multimodal models to be fully leveraged for real-world tasks. The HY-Embodied-0.5 family instantiates two primary variants: a highly efficient multimodal Mixture-of-Transformers (MoT)~\citep{mot} model (2B activated / 4B total parameters) optimized for real-time responsiveness and edge deployment, and a powerful Mixture-of-Experts (MoE)~\citep{moe} model (32B activated / 407B total parameters) engineered to tackle complex visual perception and embodied reasoning tasks. By innovating across model architecture, data curation, and training strategies, we systematically enhance the models' capabilities in both visual perception and embodied tasks. Our models achieve state-of-the-art performance across extensive perception and embodied benchmarks, with their practical effectiveness validated in downstream robotic control tasks.

To improve the visual perception and embodied understanding capabilities of the model, we propose several innovations to develop HY-Embodied-0.5. In terms of architecture, we introduce a lightweight yet powerful native-resolution Vision Transformer (ViT)~\citep{Dosovitskiyetal2020vit,Dehghanietal2023navit,tschannen2025siglip} for visual encoding, a Mixture-of-Transformers architecture to enable modality-adaptive computation and improve the model's visual modeling capacity, and incorporate visual latent tokens~\citep{zelikman2024quiet,pfau2024lets} to better connect vision and language. For data, we build high-quality perception and embodied pre-training data of over 100M training samples, covering basic perception, spatial perception, embodied perception, and reasoning and planning. By constructing real robot data and high-quality reasoning data, we improve the model's ability to solve real-world problems and complex tasks. Regarding training, we design an iterative, self-evolving post-training paradigm. We iteratively improve the thinking abilities of our model by using a small amount of cold start data, combined with iterative reinforcement learning~\citep{shao2024grpo} and rejection sampling supervised finetuning (SFT)~\citep{deepseek2025r1}. Finally, through a large-to-small on-policy distillation~\citep{agarwal2024gkd,thinkingmachines2025onpolicy} approach to transfer knowledge from the large model to the small model, we significantly improve the performance of the edge variant of our model.

Evaluation plays a central role in driving the development of HY-Embodied-0.5. To comprehensively evaluate the model's capabilities in visual perception and embodied tasks, we construct an evaluation suite comprising 22 public benchmarks, covering visual perception, spatial reasoning, and embodied understanding. Our HY-Embodied-0.5-MoT-2B achieves the best performance on 16 out of 22 benchmarks among compared generalist and specialist embodied VLMs of similar sizes. It achieves an average score of 58.0\% across all 22 benchmarks, outperforming the generalist VLM Qwen3-VL-4B~\citep{Qwen3-VL} and the specialist embodied VLM RoboBrain2.5-4B~\citep{tan2026robobrain}—both of which have larger activated parameters—by 10.2\% and 8.6\%, respectively. Notably, our embodied model also achieves comparable performance to the widely used open-source model Qwen3.5 on general VLM understanding tasks, demonstrating that our model possesses both strong generalizability and powerful embodied task capabilities. Our most powerful HY-Embodied-0.5-MoE-A32B model achieves an average score of 67.0\% across the 22 benchmarks, surpassing the frontier model Gemini 3.0 Pro (63.6\%)~\citep{gemini3}. These results strongly validate the effectiveness of our training strategy, data construction, and architectural design.

This report provides a comprehensive introduction to HY-Embodied-0.5. The rest of the report is structured as follows: Section~\ref{arch} details the model architecture and the underlying design rationale. Section~\ref{pretrain} presents the details of data construction, training recipes, and training strategies during the pre-training phase. Section~\ref{posttrain1} details the comprehensive post-training pipeline, covering data design principles and training strategies for both the SFT and RL stages, as well as the iterative post-training process and the specifics of large-to-small distillation. Section~\ref{eval} presents comprehensive quantitative and qualitative results, validating the superior performance of HY-Embodied-0.5 across various visual perception and embodied tasks. Section~\ref{vla} introduces our practices of applying the foundational VLM to downstream robot control tasks, demonstrating the strong results of our VLA model in real-world control scenarios.

%% file: sec/2-Architecture.tex
\section{Model Architecture}
\label{arch}

\begin{figure}[tb]
  \centering
  \includegraphics[width=0.95\linewidth]{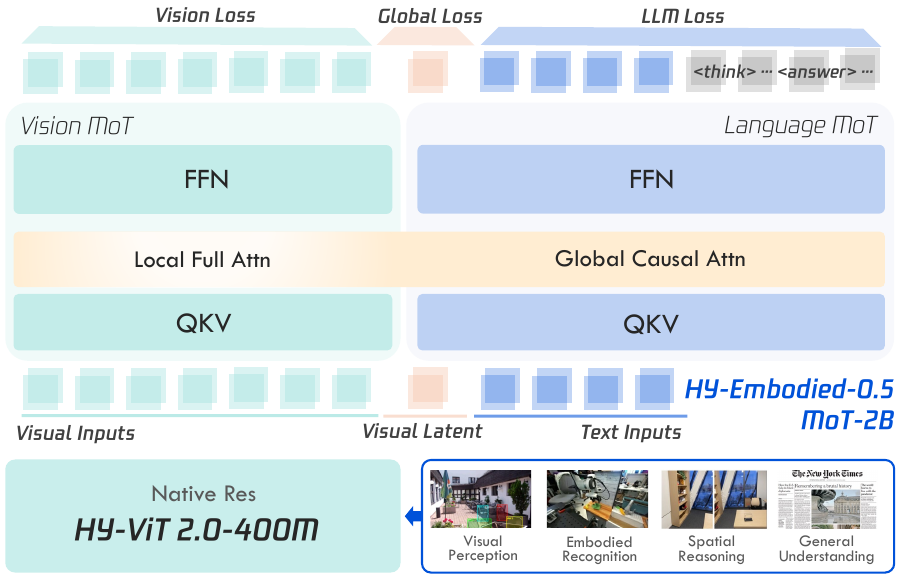}
  \caption{\textbf{HY-Embodied-0.5 Mixture-of-Transformers Architecture.} The MoT design decouples the processing of visual and textual tokens by employing modality-specific QKV and FFN layers, alongside distinct attention mechanisms. Visual latent tokens and mixed optimization loss are employed to bridge and stress the relationships between modalities during large-scale training. }
  \label{fig:2-arch}
\end{figure}

HY-Embodied-0.5 is built upon the common VLM paradigm and architecture comprising a vision encoder and a large language model. To enhance the visual perception capabilities of the model, we introduce several architectural improvements designed for the edge variant (i.e., HY-Embodied-0.5-MoT-2B) to better understand visual inputs while achieving a better balance between visual and language capabilities. Firstly, we train an efficient yet powerful native-resolution Vision Transformer (ViT)~\citep{Dosovitskiyetal2020vit,Dehghanietal2023navit} optimized for edge-device deployment. Serving as an advanced iteration of the HY-ViT series~\citep{team2025hunyuanocr}, this model inherently supports arbitrary-resolution inputs and achieves accurate, robust perception within a lightweight footprint by distilling knowledge from a larger internal model. Secondly, we adopt a Mixture-of-Transformers architecture~\citep{mot} to enable modality-adaptive computation. By introducing non-shared parameters specifically for the vision branch, we significantly boost visual performance while mitigating the degradation of language capabilities often caused by heavy visual training. We further design an independent full-attention mechanism and apply auxiliary visual supervision for the vision component to facilitate better visual modeling. Finally, inspired by recent progress in latent thinking~\citep{zelikman2024quiet,pfau2024lets} and vision registers~\citep{darcet2023registers}, we append dedicated visual latent tokens to the end of each visual input sequence. With a specifically designed supervision, these tokens further improve the models' overall perceptual capacity. The overall architecture is shown in Figure~\ref{fig:2-arch}.

\subsection{HY-ViT 2.0: Efficient Native-Resolution Visual Encoder}

The ViT model is the fundamental building block for adapting LLMs to multi-modal scenarios. It projects visual inputs into the language embedding space, enabling the LLM to seamlessly process both visual and textual inputs. The ViT model used in HY-Embodied-0.5 is an upgraded version of HY-ViT. Building upon its native support for arbitrary-resolution inputs, HY-ViT 2.0 utilizes a larger scale of pre-training data, introduces a tiny LLM to provide language supervision signals, and incorporates visual reconstruction supervision to ensure minimal information loss in the visual signals fed to the LLM. To ensure real-time performance on edge devices, we employ a 400M-parameter ViT model for HY-Embodied-0.5 and train it via distillation from a more powerful internal ViT, helping our model achieve efficient and accurate visual representations. Furthermore, we train a larger version of the ViT to generate discrete visual representations capable of both understanding and reconstruction. This representation features a codebook size of 2k and compresses every 8$\times$8 image patch into a single discrete code. We use this discrete representation to supervise the output of the model's visual tokens. Further details are provided in Section~\ref{sec:mot}.

\subsection{Modality-Adaptive Computing with Mixture-of-Transformers}
\label{sec:mot}

Adaptive computing architectures have been widely applied in Large Language Models (LLMs) and Vision-Language Models (VLMs), demonstrating an effective balance between computational efficiency and performance, typically through Mixture-of-Experts (MoE) and Mixture-of-Transformers (MoT) strategies. We incorporate the MoT architecture into our model. By introducing non-shared parameters for language and vision tokens, we improve the visual modeling capacity while mitigating the degradation of the model's inherent language capabilities caused by heavy visual training. We find this strategy especially effective for small edge models, as it doubles the inherently limited total parameter count while introducing negligible overhead to training and inference efficiency. Specifically, before multi-modal training begins, we duplicate the Feed-Forward Network (FFN) and QKV parameters of the language model, initializing these duplicated parameters with the weights of the pre-trained LLM. During the forward process, all visual tokens output by the ViT are computed using this duplicated set of parameters, while text tokens are computed using the original text-specific parameters.

\begin{wrapfigure}{r}{0.4\textwidth}
  \centering
  \vspace{-20pt}
  \includegraphics[width=\linewidth]{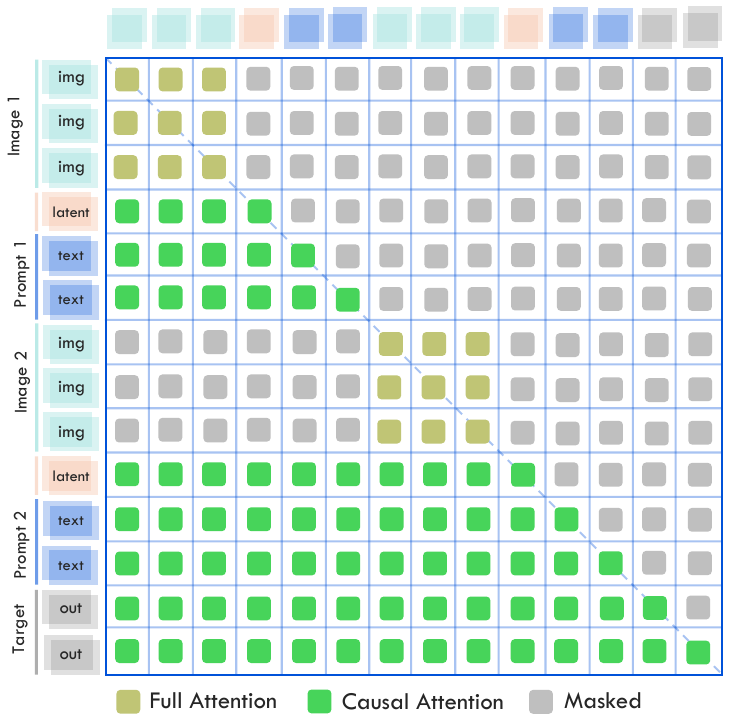}
  \caption{\textbf{Attention Computation of our Modality-Adaptive MoT.} We visualize the attention computation under actual interleaved multi-modal sequences with distinct colors.}
  \vspace{-20pt}
  \label{fig:3-mask}
\end{wrapfigure}

Beyond employing the MoT architecture, we make further improvements to better model visual inputs. As illustrated in Figure~\ref{fig:3-mask}, we design distinct attention mask patterns for visual and text tokens. Since visual data lacks the unidirectional nature characteristic of language sequences, we find that bidirectional attention is more beneficial for visual modeling, which becomes even more natural when we use the MoT architecture. Additionally, given that over half of the tokens in multi-modal data are vision tokens, we introduce a visual next-code prediction task to better optimize the vision branch in the MoT and provide stronger supervision signals. Specifically, using the discrete visual representations from the larger ViT as supervision, we apply an MLP module to the LLM output features of the vision branch to predict the discrete code of the next patch (see \emph{Vision Loss} in Figure~\ref{fig:2-arch}). These designs enable the MoT to achieve better visual modeling, effectively improving the model's overall visual capabilities and performance on fine-grained perception tasks.


\subsection{Visual Latent Tokens Connecting Vision and Language}

Inspired by recent progress in latent thinking~\citep{zelikman2024quiet,pfau2024lets} and vision registers~\citep{darcet2023registers}, we find that appending a learnable visual latent token to the end of each visual element (e.g., an image or a video frame) is beneficial for improving the capabilities of small VLMs. Furthermore, during the pre-training phase, we use the global features from a large ViT to supervise the output features of this token, which further improves model performance (see \emph{Global Loss} in Figure~\ref{fig:2-arch}). We observe that this visual latent token effectively connects visual and textual content. A more intuitive understanding of its function can be seen from the visualization in Figure~\ref{fig:11-visual_latent_token}.

%% file: sec/3-Pretrain.tex
\section{Pre-training}
\label{pretrain}

Building upon the HY large language model (Hunyuan-1.8B, \cite{tencent2025hunyuan18b}), our training pipeline embeds physical world understanding from the earliest phases of training. Specifically, during the initial large-scale pre-training stage, we introduce a diverse and extensive corpus of visual perception data—spanning 2D and 3D grounding, depth estimation, and image segmentation. This early integration fundamentally enhances the model's capacity to perceive and interpret complex physical environments. Following this, a targeted mid-training stage aligns the model's capabilities with downstream embodied requirements. By blending rich embodied and spatial datasets with general-domain data, we effectively enhance the model's spatial cognition and complex reasoning capabilities for real-world agentic applications.

\subsection{Pre-training Data}

We compile a diverse and high-quality set of vision-language data to formulate our pre-training and mid-training mixtures. Specifically, the data composition integrates low-level visual perception data with dedicated datasets formulated for embodied tasks and spatial cognition. To construct the overall training corpus, these specialized, domain-specific data sources are directly combined with large-scale general understanding data.

\subsubsection{Visual Perception Data}

\paragrapha{Omni-Detection. }We curate an Omni-Detection dataset comprising both 2D and 3D detection data to strengthen the model’s grounding and object recognition capabilities. Source images are drawn from large-scale datasets, including OpenImages~\citep{kuznetsova2020open}, Objects365~\citep{shao2019objects365}, RefCOCO~\citep{yu2016modeling}, SA-1B~\citep{kirillov2023segment}, etc. For samples with high-quality annotations, we directly convert the existing labels into a unified detection format. For unlabeled data or those with low-quality annotations, we employ an automated labeling pipeline: we first utilize a VLM to identify objects, then combine SAM~\citep{carion2025sam3segmentconcepts} with VLM grounding to determine their coordinates. A stronger VLM teacher is subsequently deployed to verify the accuracy of these generated annotations. The detection tasks encompass object tagging and the prediction of 2D and 3D bounding boxes. We obtain 62M Omni-Detection data in total. In our implementation, all coordinates are normalized to integers ranging from 0 to 1000 and represented in a fixed output format.

\paragrapha{Depth Estimation. }Depth estimation, encompassing both absolute and relative depth, serves as a critical channel for embodied VLMs to perceive the physical environment. We derive sensor-based ground truth from large-scale indoor and outdoor 3D datasets, alongside autonomous driving corpora, to construct question-answering pairs based on specific image coordinates. To ensure data quality, our point-sampling strategy explicitly excludes pixels located on object boundaries, at infinity, or within physically inconsistent regions. To facilitate effective data fusion across diverse sources, we normalize the camera focal lengths across all images, thereby standardizing the scale of depth measurements. In addition to absolute metric depth, we generate a substantial volume of relative depth data based on real-world distances to enhance the model’s comprehensive spatial understanding. This process results in a specialized dataset comprising approximately 36M samples.

\paragrapha{Segmentation. }For semantic segmentation, we source high-resolution, high-quality segmentation maps from the SA-1B dataset~\citep{kirillov2023segment}. Given the dense and highly detailed nature of SA-1B annotations, we apply a filtering process to remove excessively small, disproportionately large, and highly fragmented object masks. Upon obtaining the refined binary mask matrices, we adopt the methodology established by PaliGemma~\citep{beyer2024paligemma}. Specifically, we expand our tokenizer vocabulary to encode these masks, converting them into structured question-answering pairs formatted for VLM prediction. This pipeline yields approximately 5M segmentation samples, designed to enhance the model's fine-grained visual perception and edge-awareness capabilities.

\paragrapha{Pointing and Counting. }Object pointing and counting are notoriously challenging tasks for VLMs, frequently leading to enumeration errors and spatial hallucinations. However, precise point-level perception is essential for fine-grained embodied manipulation. To explicitly reinforce the model's comprehensive object comprehension, we formulate a specialized, high-difficulty pointing and counting dataset. Specifically, we source ground-truth point annotations from open-source datasets such as Pixmo-Points~\citep{deitke2025molmo}. To ensure sufficient task complexity, we deliberately filter and select scenes containing a high density of target objects from our broader detection corpora. We obtain approximately 11M object-counting and pointing data for training.

\begin{figure}[tb]
  \centering
  \includegraphics[width=\linewidth]{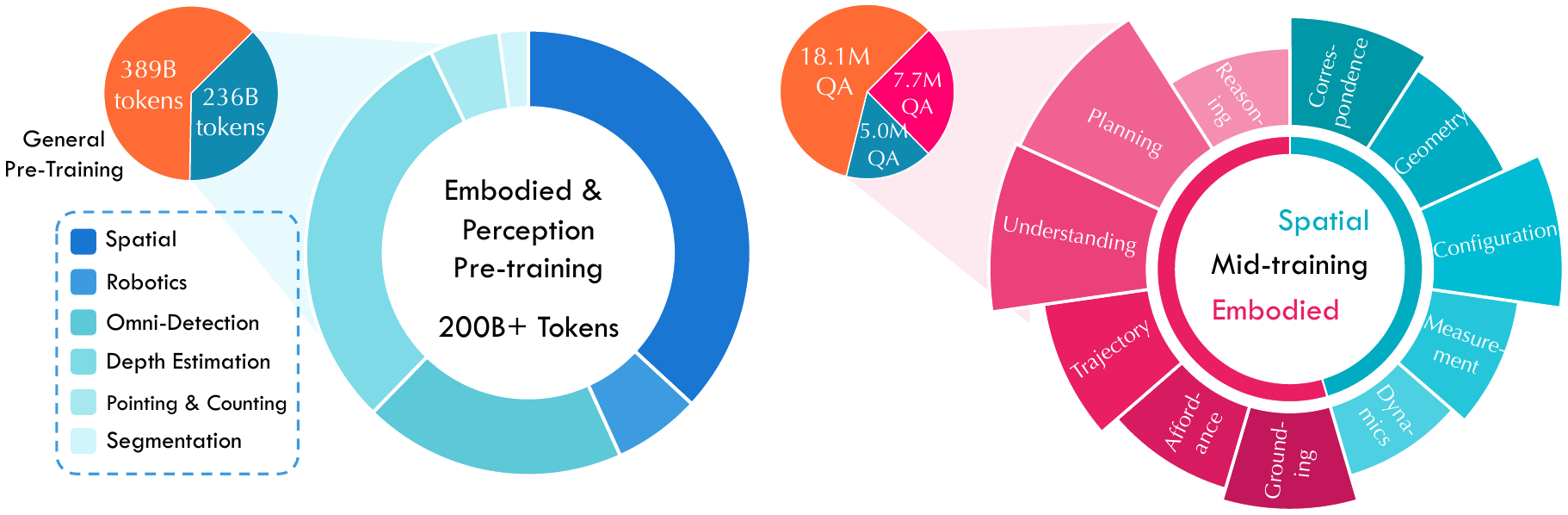}
  \caption{\textbf{Data Distribution for Pre-training and Mid-training Stages.} We conduct large-scale embodied pre-training and mid-training to establish foundational and advanced physical-world competencies. The pre-training mixture encompasses over 200B tokens based on spatial, robotics, and visual perception tasks. The mid-training stage leverages over 12M high-quality QA pairs for complex real-world execution based on diverse spatial and embodied domains.}
  \label{fig:4-pretrain}
\end{figure}

\subsubsection{Embodied-Centric Data}

To construct the embodied dataset, we aggregate open-source annotated data alongside ego-view robotics manipulation sequences recorded in real-world environments. To systematically address the operational requirements of physical agents, we organize our data into a three-tiered hierarchy: embodied perception, semantic understanding, and high-level planning and manipulation. The perception tier establishes foundational spatial and physical awareness; the semantic tier bridges visual inputs with contextual reasoning; and the planning tier provides supervision for sequential decision-making and action execution.

\paragrapha{Grounding. }Visual grounding provides the foundational spatial guidance required for embodied execution. Building upon the large-scale perception pre-training, we incorporate scenarios directly aligned with physical manipulation. This dataset is compiled from open-source datasets, including Molmo~\citep{deitke2025molmo}, RoboPoint~\citep{yuan2024robopoint}, and RefSpatial~\citep{zhou2025roborefer}, alongside our in-house annotations. The defined tasks encompass point-level object localization, bounding box prediction, and referring expression comprehension. During the data filtering and annotation process, we isolate elements critical to embodied operations, such as target interactive objects and the robotic manipulators themselves. This targeted selection explicitly reinforces the model's spatial recognition within operational environments.

\paragrapha{Affordance. }Affordance prediction integrates visual grounding with user instructions, demanding a higher level of task comprehension. We source training data from established affordance benchmarks, including RoboAfford~\citep{hao2025roboafford++} and ShareRobot~\citep{tan2026robobrain}. Additionally, we repurpose a subset of our existing embodied grounding data. Specifically, we employ a VLM to generate contextually appropriate user instructions, directly pairing the original spatial grounding annotations with these generated operational commands.

\paragrapha{Trajectory. }Trajectory prediction is essential for internal planning in embodied tasks; however, the utility of such data is constrained by factors like waypoint density and spatial accuracy. We source annotated trajectory data from open-source datasets, including MolmoAct~\citep{lee2025molmoact}, ShareRobot~\citep{tan2026robobrain}, and FSD~\citep{yuan2025seeing}. Furthermore, we extract actual motion trajectories from large-scale embodied manipulation video clips~\citep{wu2024robomind,o2024open,khazatsky2024droid} by employing the tracking model cotracker3~\citep{karaev2025cotracker3} to trace the position of the robotic arm or agent. For these extracted trajectories, we retain the first frame of the video as the visual input for the question-answering pair. The extracted sequences are then downsampled to a maximum of 15 waypoints and visually plotted onto the image. Finally, we deploy a stronger VLM judge to evaluate the accuracy of these plotted trajectories and filter the data accordingly.

\paragrapha{Understanding. }Embodied understanding represents a synthesis of multi-level VLM capabilities, encompassing foundational spatial cognition, task state evaluation, planning strategy formulation, and the interpretation of in-image annotations. We source a substantial volume of question-answering data from open-source datasets, including Robo2VLM~\citep{chen2025robo2vlm}, RoboVQA~\citep{sermanet2024robovqa}, RoboRefit~\citep{lu2023vl}, and RoboInter-VQA~\citep{li2026robointer}. To construct the final understanding dataset, we filter these raw QA pairs based strictly on data quality and annotation accuracy.

\paragrapha{Planning. }Embodied planning requires the model to assess the current execution state and comprehend the target task objective. To construct this dataset, we utilize a VLM to annotate the primary tasks within robotic manipulation video clips~\citep{wu2024robomind,bu2025agibot_iros,wu2025robocoin}. We then temporally segment these videos to extract ground-truth labels for subsequent actions. The resulting segments are formatted into query-response pairs that prompt the model to predict future action sequences. Additionally, we explicitly define task constraints within the user instructions to enhance the model's instruction-following capabilities. Finally, we supplement these generated sequences with open-source planning question-answering pairs sourced from datasets such as RoboVQA~\citep{sermanet2024robovqa} and RoboInter~\citep{li2026robointer}.

\paragrapha{Reasoning. }To extend the model's capabilities beyond standard operational instructions, we construct a complex, in-house reasoning dataset situated in real-world embodied environments. This corpus specifically targets scenarios demanding long-horizon reasoning. The problem scope encompasses action sequencing, multi-image action comprehension, future state prediction, visual puzzle resolution, and intuitive physics reasoning.

\subsubsection{Spatial-Centric Data}

Spatial-centric data focuses on understanding and reasoning about three-dimensional environments from visual observations. Unlike embodied-centric data that emphasizes agent-environment interactions, spatial-centric data targets the fundamental capabilities of perceiving geometric structures, establishing visual correspondences, and reasoning about spatial relationships. We categorize spatial-centric data into five types: Correspondence, Geometry, Configuration, Measurement, and Dynamics. Raw data is sourced from ScanNet~\citep{dai2017scannet}, ScanNet++~\citep{yeshwanth2023scannet++}, ARKitScenes~\citep{baruch2021arkitscenes} and our self-collected data.

\paragrapha{Correspondence.} Correspondence data establishes associations between visual elements across different viewpoints, frames, or representational spaces (e.g., 2D to 3D). This capability is fundamental for multi-view understanding and serves as the foundation for downstream spatial reasoning tasks. Our correspondence data includes two primary forms: (1) cross-frame point matching that identifies corresponding points between temporally adjacent frames, formulated as both coordinate-based and visual marker-based question-answering pairs; and (2) 2D-3D instance mapping that links 2D bounding boxes in image space to 3D instance identifiers in the scene representation. We leverage posed RGB-D sequences, where camera intrinsics and extrinsics enable precise projection between coordinate systems. The data generation pipeline first computes visibility information for sampled frames, then generates QA pairs that query point correspondences using either explicit coordinates or visual dot markers overlaid on images.

\paragrapha{Geometry.} Geometry data captures the three-dimensional structure of scenes, including depth relationships and spatial extent. We focus on depth perception tasks that require understanding relative and absolute distances from the camera viewpoint. The geometry data encompasses: (1) depth estimation, where models predict the depth of specific points indicated by coordinates or visual markers; and (2) depth comparison, where models determine which of two indicated points is closer to or farther from the camera. Both coordinate-based and visual dot-based formulations are included to evaluate different input modalities. The pipeline samples point pairs with sufficient depth disparity (typically >0.3m) to ensure unambiguous annotations.

\paragrapha{Configuration.} Configuration data addresses the spatial arrangement and relationships between objects within a scene. This category covers static spatial understanding without explicit metric measurements. We generate four types of configuration QA pairs: (1) object counting that queries the number of instances for specific categories or combinations of categories; (2) relative distance identification that determines which object among a candidate set is closest to a reference object; (3) relative direction determination that identifies the directional relationship (left, right, front, back) of a target object relative to an observer's position and facing direction; and (4) distance ranking that orders multiple objects by their proximity to a reference. The data is derived from 3D bounding box annotations and instance segmentation. For relative direction tasks, we compute angles on the 2D ground plane between the observer's forward vector and the query vector, with filtering to exclude ambiguous cases where angular differences are insufficient.

\paragrapha{Measurement.} Measurement data requires precise metric estimation of spatial quantities. Unlike configuration data that focuses on relative relationships, measurement tasks demand numerical outputs in physical units. We include three measurement types: (1) object size estimation that predicts the longest dimension of an object in centimeters, derived from the maximum axis length of oriented 3D bounding boxes; (2) absolute distance computation that measures the Euclidean distance between two objects in meters, calculated as the minimum distance between their 3D bounding boxes; and (3) room size estimation that predicts the total floor area in square meters. These tasks are generated from scenes with calibrated 3D reconstructions, ensuring metric accuracy. We apply filtering to exclude trivially close object pairs ($<$0.2m) for distance tasks and restrict size queries to objects with unique instances to avoid ambiguity.

\paragrapha{Dynamics.} Dynamics data captures motion and temporal changes in spatial environments, including both camera ego-motion and object movement. For camera dynamics, we generate data that describes the spatial transformation between frames, including relative rotation and translation patterns. The pipeline first computes frame-to-frame geometric relations and stores them in a structured format, then generates QA pairs querying camera movement characteristics. For object dynamics, we leverage dense 3D point tracks across video sequences, with annotations containing 3D coordinates, visibility flags, and camera extrinsics. QA pairs query object movement patterns using either coordinate-based or visual marker-based formulations.

\subsubsection{General Understanding Data}

We incorporate a substantial volume of in-house general VLM data to establish the model's foundational reasoning and comprehension capabilities. This diverse corpus is systematically categorized to target several core domains: general semantics (e.g., image captioning and world knowledge), STEM proficiency (e.g., mathematics, coding, and scientific reasoning), fine-grained visual parsing (e.g., document understanding, charts, and OCR), complex problem-solving (e.g., logical reasoning, multi-round dialogues, multi-image contexts, and complex instruction following), and agentic operations (e.g., GUI navigation). To optimize the learning curriculum, we partition these general datasets into two distinct subsets based on their scale and annotation quality, allocating them to the pre-training and mid-training stages, respectively. Throughout both training phases, these broad-domain datasets are jointly trained with the specialized embodied corpora, ensuring a robust baseline performance alongside advanced physical-world competencies.

\begin{figure}[tb]
  \centering
  \includegraphics[width=\linewidth]{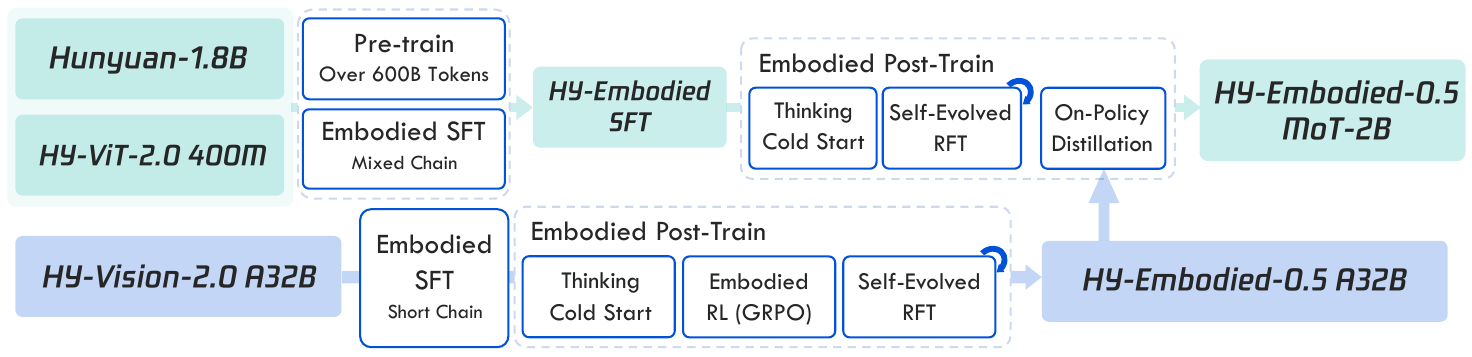}
  \caption{\textbf{Training Pipeline for HY-Embodied-0.5 Series.} Large-scale pre-training establishes the models' foundational multi-modal representations and robust spatial-embodied perception. The subsequent Embodied Post-training phase explicitly enhances complex reasoning capabilities through iterative self-evolution and reinforcement learning. Finally, we employ on-policy distillation to effectively transfer the knowledge from large variants to edge deployment.}
  \label{fig:5-pipeline}
\end{figure}

\subsection{Training Recipe}

As illustrated in Figure~\ref{fig:4-pretrain}, our training recipe is structured into two sequential stages. The first stage consists of large-scale pre-training, where the model is optimized over a massive multimodal corpus comprising more than 600B tokens to establish foundational visual-linguistic alignment. Following this phase, the model undergoes a dedicated mid-training stage. In this second phase, training is conducted on a carefully curated mixture of approximately 25M data samples, seamlessly integrating the aforementioned embodied, spatial, and general understanding datasets.

\paragrapha{Pre-training. } In this stage, the training corpus comprises 389 billion tokens of general understanding data and 236 billion tokens of embodied and perception data. Within the latter, spatial and robotics data account for 43\%, and visual perception data make up the rest. We allocate large-scale data with homogeneous patterns to this pre-training phase, reserving more fine-grained data for the mid-training stage. We set the base learning rate to 5e-5, the ViT learning rate to 5e-6, the weight decay to 1e-4, and the global batch size to 256. Training samples are packed to a maximum context length of 32k tokens based on the length of each question-answering pair. The parameters of the ViT, the MoT module, and the latent visual tokens are all trainable. The gradients for the ViT and visual tokens are updated once every five steps.

\paragrapha{Embodied-Spatial Mid-training. }In the Embodied-Spatial Mid-training stage, we introduce higher-quality and more complex embodied and spatial data, encompassing approximately 30 million instances. We mix the general understanding, embodied, and spatial data at a ratio of 12:5:3, and unify all prompt templates and coordinate formats. We apply variant-specific data strategies during this phase: for the MoT-2B model, we utilize a mixture of long and short reasoning chains, differentiated via $\backslash think$ and $\backslash no\_think$ tokens following Qwen3-VL; conversely, for the MoE-32B model, we exclusively employ short-chain data to concentrate on embodied fine-tuning. During training, we retain the sequence packing method and initial learning rate from the pre-training stage, while introducing a cosine learning rate decay. We freeze all ViT parameters and exclusively update the HY-Embodied-0.5 modules.

\subsection{Training Strategy}

Based on our Mixture-of-Transformers (MoT) design, we adopt a vision loss, a global loss, and a standard LLM loss to respective supervise the visual tokens, latent tokens, and language tokens. For the visual next-code prediction task, we apply a cross-entropy loss over the predicted logits from the vision branch. Let $N_v$ denote the number of visual tokens, $p_i$ be the predicted probability distribution for the $i$-th token, and $z_i$ be the target discrete code generated by the teacher ViT. The vision loss is formulated as:
$$\mathcal{L}_{\text{vision}} = -\frac{1}{N_v} \sum_{i=1}^{N_v} \log p_i(z_i)$$

To explicitly align the visual latent token with the overarching image semantics, we compute the negative cosine similarity between the mapped hidden states of the latent token ($f_{\text{latent}}$) and the global CLS feature extracted from the teacher ViT ($f_{\text{teacher}}$). The global loss is defined as:

$$\mathcal{L}_{\text{global}} = - \frac{f_{\text{latent}}^\top f_{\text{teacher}}}{\|f_{\text{latent}}\| \|f_{\text{teacher}}\|}$$

During the large-scale pre-training phase, the model is jointly optimized using the summation of these three objectives: $\mathcal{L}_{\text{total}} = \mathcal{L}_{\text{llm}} + \mathcal{L}_{\text{vision}} + \mathcal{L}_{\text{global}}$. In the subsequent mid-training and all fine-tuning stages, we discard the vision and global supervision signals, exclusively optimizing the standard autoregressive language loss ($\mathcal{L}_{\text{llm}}$).

%% file: sec/4-Posttrain1.tex
\section{Post-training}
\label{posttrain1}

\subsection{Supervised Fine-tuning}

\subsubsection{Data Construction}

In the supervised fine-tuning (SFT) stage, we focus on reinforcing the models' long-chain reasoning capabilities. We sample a subset of high-complexity, multi-step problems from the aforementioned spatial, embodied, and general data sources, together with more in-house reasoning data. For these instances, we construct Chain-of-Thought (CoT)~\citep{wei2022chain} trajectories via a human-model collaborative pipeline. The generated CoTs are subsequently evaluated by an LLM across multiple dimensions, including reasoning quality, logical correctness, and sequence repetition. We additionally verify the exact match accuracy of the final deduced answers. This pipeline yields approximately 100k cold-start CoT instances, which are utilized to train the MoT-2B and MoE-A32B variants.

\subsubsection{Training Recipe}

During the cold-start SFT phase, we continue to optimize the models using the standard cross-entropy loss. However, unlike the pre-training and mid-training stages, we explicitly disable sequence packing. Each training sample is processed individually to isolate and emphasize the independent reasoning chain of each data entry. We maintain the base learning rate at 5e-5 throughout this training process.

\subsection{Reinforcement Learning}

\subsubsection{Data Construction}

For reinforcement learning, we dynamically construct the training data according to the current model capability, rather than relying on a fixed dataset. We maintain a large candidate pool covering diverse embodied capabilities, and in each RL round use the latest model to perform multi-sample evaluation on this pool. Samples that are solved correctly in all attempts are discarded as overly easy, while samples that fail in all attempts are removed as overly difficult. We retain only samples with partial success, as these examples lie near the model's current capability frontier and typically provide the most informative learning signals for policy improvement.

To avoid over-optimizing RL toward a narrow subset of embodied capabilities, we further balance the selected data across different capability dimensions, including perception, prediction, interaction, and planning. Each RL stage is trained on a newly constructed set of 50K samples. As the model improves, this procedure continuously refreshes the effective training distribution, forming a simple capability-adaptive curriculum that stabilizes optimization and supports sustained gains in embodied reasoning.

\begin{figure}[tb]
  \centering
  \includegraphics[width=\linewidth]{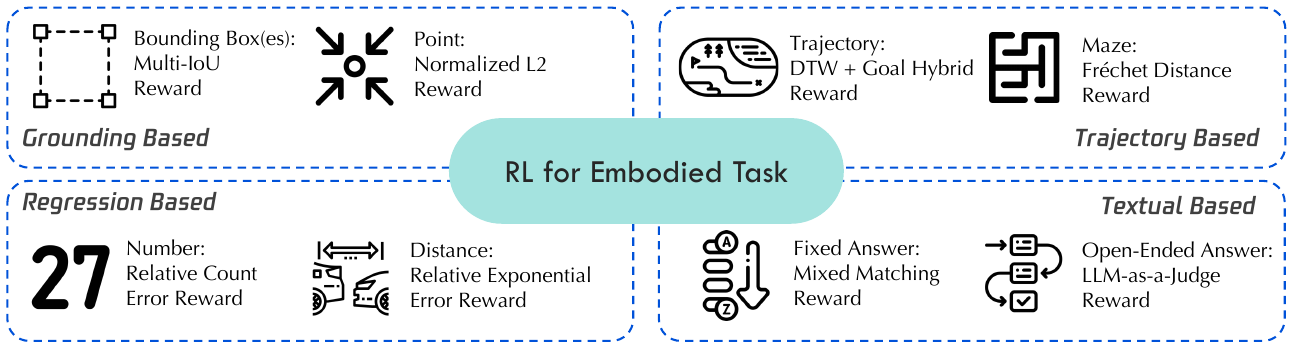}
  \caption{\textbf{Reward Designs for Embodied Reinforcement Learning. }To accommodate diverse embodied tasks during RL optimization, we systematically formulate reward functions into four categories: Grounding-Based for spatial localization, Regression-Based for numerical estimation, Trajectory-Based for motion and planning, and Textual-Based for general and semantic reasoning.}
  \label{fig:6-rl}
\end{figure}

\subsubsection{Reward Designs}

A key challenge of RL for embodied models is that the target outputs are highly heterogeneous, spanning geometric grounding, trajectory prediction, discrete decisions, continuous estimation, and open-ended reasoning. A single uniform reward is therefore inadequate. In our RL stage, we adopt a task-aware reward design, where each sample is assigned a reward function according to the structure of its target output:
\begin{equation}
    r = R_t(y, y^\star) \in [0,1].
\end{equation}
Our principle is to use deterministic and structure-aware rewards whenever the target admits reliable parsing, and to resort to an LLM judge only when deterministic evaluation is insufficient.

For tasks with explicit geometric structure, we use dense rewards based on geometric similarity rather than exact match. Concretely, grounding tasks are evaluated by overlap- or distance-based measures such as IoU, Hungarian-matched IoU, normalized point distance, and Chamfer distance, which provide graded supervision for localization and fine-grained perception. For trajectory prediction and planning tasks, we similarly use path-aware rewards based on sequence or curve similarity, such as DTW- and Fr\'echet-distance-based scores, optionally combined with endpoint consistency terms. These rewards are important for embodied settings, where partial spatial or temporal correctness should be distinguished from complete failure.

For outputs with discrete or strongly constrained formats, we use lighter-weight exact-match-style rewards. This includes multiple-choice prediction, binary judgment, counting, and other structured-answer tasks. When the target is sequential but discrete, such as sorting or ordering, we instead use partial-credit rewards based on sequence similarity, e.g., normalized longest common subsequence. For continuous estimation tasks, we adopt regression-style rewards that decay smoothly with relative error, which provide more informative signals than hard-threshold accuracy.

Finally, for open-ended embodied reasoning tasks whose correctness cannot be robustly determined by rules alone, we use an LLM-based judge as a fallback:
\begin{equation}
    r_{\text{free}} = J(q, y, y^\star),
\end{equation}
where $q$ is the input, $y$ is the model response, and $y^\star$ is the reference answer. This fallback extends the reward framework to free-form reasoning tasks while preserving deterministic scoring whenever possible.

Overall, our reward design follows a simple principle: reward structure should match output structure. Dense geometric and regression-style rewards are used where partial correctness is meaningful, exact matching is used where the answer space is unambiguous, and LLM-based judgment is reserved for genuinely open-ended cases. We find this hybrid design important for stabilizing RL over diverse embodied capabilities.

\subsubsection{Training Recipe}

We optimize the model in the RL stage with a GRPO-based objective~\citep{shao2024grpo}. For each multimodal input $x=(I,Q)$, we sample a group of $G$ responses from the current policy $\pi_{\theta_{\mathrm{old}}}$ and score them with the task-aware reward in Section~4.2.2. Let the resulting rewards be $\{r_1,\dots,r_G\}$. We compute group-relative advantages by normalizing rewards within each sampled group,
\begin{equation}
    A_i = \frac{r_i-\mu(\mathbf{r})}{\sigma(\mathbf{r})},
    \qquad
    \mathbf{r}=\{r_1,\dots,r_G\},
\end{equation}
and share the same advantage across all tokens in the corresponding rollout. This relative normalization is particularly suitable for embodied RL, where tasks are highly heterogeneous and raw reward scales are not directly comparable across samples.

The policy is then updated with a clipped policy-ratio objective:
\begin{small}
\begin{align}
\label{eq:hy_embodied_grpo_compact}
\mathcal{L}_{\mathrm{RL}}(x)
=
-\frac{1}{\sum_{i=1}^{G}|y_i|}
\sum_{i=1}^{G}\sum_{t=1}^{|y_i|}
\min\!\Big(
\rho_{i,t}A_i,\,
\mathrm{clip}(\rho_{i,t}, 1-\epsilon_{\mathrm{low}}, 1+\epsilon_{\mathrm{high}})A_i
\Big),
\end{align}
\end{small}
where
\begin{equation}
    \rho_{i,t}
    =
    \frac{\pi_{\theta}(y_{i,t}\mid x,y_{i,<t})}
    {\pi_{\theta_{\mathrm{old}}}(y_{i,t}\mid x,y_{i,<t})}.
\end{equation}
In practice, we use a group size of $G=16$ and adopt an essentially on-policy update scheme by matching the PPO~\citep{schulman2017ppo} mini-batch size to the rollout batch size.

To stabilize training, we further incorporate several practical controls. We mask groups with zero reward variance, since they do not provide meaningful relative learning signals. We also apply quality control to overlong and repetitive responses, and use additional length-related shaping for selected subjective tasks. Moreover, we adopt asymmetric clipping with an effective importance-ratio range of $[0.8,\,1.35]$, which we find more stable than a symmetric clipping rule in long-chain multimodal RL.

In all RL experiments, we use a maximum prompt length of 16{,}384 tokens and a maximum response length of 16{,}384 tokens. Rollouts are sampled with temperature $1.0$, top-$p=1.0$, and top-$k=-1$. The training batch size is 128, the learning rate is $8\times10^{-7}$, and each RL stage is run for 5 epochs. We also enable standard memory-efficient training techniques such as gradient checkpointing and parameter/optimizer offloading to support stable optimization of the large embodied model.

%% file: sec/5-Posttrain2.tex

\subsection{Evolving Deep Thinking with Iterative Training}

While RL directly improves task reward, it does not necessarily guarantee high-quality reasoning traces. In embodied tasks, correct answers may arise from very different internal processes, ranging from coherent spatial reasoning to unstable shortcuts. To further improve the depth and consistency of reasoning, we introduce an iterative self-evolving training paradigm based on rejection sampling fine-tuning (RFT).

Starting from the latest model checkpoint after RL, we perform multi-sample rollout on a curated data pool and evaluate the sampled responses offline using criteria aligned with the reward functions in RL. We then retain only samples that are solved correctly in some, but not all, rollouts. This filtering removes examples that are already saturated as well as examples that remain out of reach, and concentrates on the model's current learnable frontier. Among these retained samples, we further score the quality of the reasoning traces with a stronger teacher model and keep only those whose thinking quality exceeds a predefined threshold. In practice, this process filters approximately 1M candidate examples into around 300K high-quality traces for the subsequent SFT stage.

The role of RFT is complementary to RL. RL is effective for exploration, as it helps the model discover better behaviors through reward-driven search, but its supervision is indirect and relative. RFT instead converts these discoveries into explicit positive supervision by selecting high-quality successful traces and training the model to reproduce them. In this sense, RL expands the capability frontier, while RFT consolidates the best newly discovered reasoning patterns into more stable behavior.

We therefore alternate RL and RFT throughout post-training. In each cycle, RL improves the model through online optimization, and RFT distills the resulting high-quality reasoning traces through supervised refinement. This iterative process gradually transforms occasional success into reliable capability, and we find it particularly effective for cultivating deep thinking in embodied models.

\subsection{Large-to-Small On-Policy Distillation}

Although the large HY-Embodied-0.5 model exhibits substantially stronger embodied reasoning ability, our practical deployment target is the compact model. We therefore introduce a large-to-small on-policy distillation stage to transfer the teacher's reasoning behavior into the student. The goal of this stage is not merely model compression, but preserving as much of the teacher's embodied competence and thinking style as possible under a much smaller capacity budget.

The key observation is that reasoning ability is not only reflected in final outputs, but also in the token-level continuation distribution along the generation process. Standard offline distillation on teacher-generated responses is therefore insufficient, since it only exposes the student to teacher trajectories and does not supervise the student on its own decoding states. To address this, we adopt an on-policy distillation strategy: the student first rolls out its own response
\begin{equation}
    y=(y_1,\dots,y_T)\sim \pi_s(\cdot\mid x),
\end{equation}
and the teacher is then applied under teacher forcing on the same student-generated prefixes. Let $\pi_t(\cdot\mid x,y_{<t})$ and $\pi_s(\cdot\mid x,y_{<t})$ denote the teacher and student next-token distributions at step $t$, respectively. We optimize the student by minimizing
\begin{equation}
\label{eq:opd_compact}
    \mathcal{L}_{\mathrm{OPD}}
    =
    \mathbb{E}_{x,\,y\sim\pi_s(\cdot\mid x)}
    \left[
    \frac{1}{|y|}
    \sum_{t=1}^{|y|}
    \mathrm{KL}\!\left(
        \pi_t(\cdot\mid x,y_{<t})
        \,\|\, 
        \pi_s(\cdot\mid x,y_{<t})
    \right)
    \right].
\end{equation}

This design allows the student to learn from the teacher precisely on the states induced by its own policy, where its errors actually occur. Compared with conventional offline distillation, it substantially reduces the mismatch between training and inference, and transfers a richer signal than final-answer imitation alone. In our setting, this is particularly important because the capabilities acquired by the large model through RL and RFT are distributed across the entire reasoning process rather than concentrated only in the final answer.

OPD also fits naturally into our post-training pipeline. RL expands the capability frontier of the large model, RFT consolidates newly discovered high-quality reasoning traces, and OPD then transfers these refined behaviors into the compact model. In this sense, OPD serves as the final bridge from capability discovery in the large model to capability deployment in the small model, enabling the released compact model to inherit a substantial portion of the teacher's embodied reasoning ability.

%% file: sec/6-Evaluation.tex
\section{Evaluation}
\label{eval}

\subsection{Results of HY-Embodied-0.5 MoT-2B}

\input{tab/embodied_table_hymot2b}
\paragrapha{Evaluation Settings.} We evaluate HY-Embodied-0.5 MoT-2B on a comprehensive suite of 22 benchmarks covering visual perception, embodied understanding, and spatial understanding. 

We evaluate HY-Embodied-0.5 MoT-2B on a comprehensive suite of 22 benchmarks covering visual perception, embodied understanding, and spatial understanding. To assess foundational visual and multimodal capabilities, we utilize CV-Bench~\citep{tong2024cambrian} and DA-2K~\citep{yang2024depth}. Moving beyond basic perception, the model’s physical and geometric reasoning is tested through benchmarks focusing on 3D spatial comprehension and multi-view geometry, including 3DSRBench~\citep{ma20253dsrbench}, EmbSpatial-Bench~\citep{du2024embspatial}, RoboSpatial-Home~\citep{song2025robospatial}, All-Angles Bench~\citep{yeh2026seeing}, MindCube~\citep{yin2025mindcube}, and MMSI-Bench~\citep{yang2025mmsi}. We further evaluate its situated environmental awareness and spatial grounding using RefSpatial-Bench~\citep{zhou2025roborefer}, SAT~\citep{ray2024sat}, SIBench-mini~\citep{yu2025far}, SITE-Bench~\citep{wang2025site}, ViewSpatial~\citep{li2025viewspatial}, VSIBench~\citep{yang2025thinking}, and Where2Place~\citep{yuan2024robopoint}. Finally, to measure embodied agency, encompassing affordance recognition, trajectory prediction, and complex task planning, the model is evaluated on ERQA~\citep{team2025gemini} RoboBench-MCQ~\citep{luo2025robobench}, RoboBench-Planning~\citep{luo2025robobench}, ShareRobot~\citep{ji2025robobrain}-Affordance and Trajectory, and Ego-Plan2~\citep{qiu2024egoplan}.

Unless otherwise specified, we report the micro-average score over all evaluation samples. For several benchmarks with task-specific protocols, we follow their corresponding metrics: 3DSRBench and SAT are evaluated using circular accuracy, ShareRobot-Bench-Affordance is evaluated using mIoU, and ShareRobot-Bench-Trajectory is evaluated using 1-DFD, where DFD denotes Dynamic Fr\'echet Distance. Since lower DFD indicates better trajectory similarity, we report $1-\mathrm{DFD}$ so that higher values consistently indicate better performance across benchmarks. We use the same evaluation setting for the A32B model, so that results are directly comparable across model scales.

\paragrapha{Baselines and Reporting Protocol.} We compare HY-Embodied-0.5 MoT-2B with representative generalist and specialist embodied VLMs, including Qwen3-VL~\citep{Qwen3-VL}, RoboBrain~\citep{tan2026robobrain}, and MiMo-Embodied~\citep{hao2025mimoembodiedxembodiedfoundationmodel}. For the Qwen family, we use Qwen3-VL as the main baseline rather than Qwen3.5-VL. In our evaluation setting, we observe that Qwen3.5-VL often produces excessively repetitive outputs, which can lead to overlong thinking sequences and significantly degraded evaluation results. To reduce the impact of such mode-specific instability and make the comparison more robust, for all baseline models we report the better result between thinking and non-thinking modes. In contrast, for HY-Embodied-0.5 MoT-2B we report the result in thinking mode. This makes the comparison conservative with respect to our model.

\paragrapha{Main Results.} Table~\ref{tab:hymot2b_main_table} summarizes the benchmark results of HY-Embodied-0.5 MoT-2B. Overall, our model achieves the best performance on 16 out of 22 benchmarks and ranks second on 4 additional benchmarks, showing strong and consistent performance across a broad range of embodied tasks.

Across the three evaluation categories, HY-Embodied-0.5 MoT-2B demonstrates a well-balanced capability profile. It achieves leading results on visual perception benchmarks, indicating that the proposed visual architecture provides a strong foundation for downstream embodied reasoning. On embodied understanding tasks, the model also performs competitively and shows clear strengths in perception, grounding, and structured decision-making, while remaining competitive on planning- and trajectory-intensive benchmarks. Its most significant advantage appears on spatial understanding benchmarks, where it consistently outperforms competing models on the majority of tasks. This strong spatial performance suggests that HY-Embodied-0.5 MoT-2B has developed particularly effective fine-grained spatial reasoning ability, which is a key requirement for real-world embodied agents.

Another notable observation is that HY-Embodied-0.5 MoT-2B remains highly competitive despite its compact size. Compared with larger baselines, our 2B model still achieves superior performance on most benchmarks, suggesting that the gains do not come from scale alone, but also from our embodied-centric design in architecture, data construction, and post-training. Overall, these results show that HY-Embodied-0.5 MoT-2B achieves a strong balance between compact model size and embodied capability, making it a strong edge model for real-world agent deployment.

\paragrapha{Results on General Benchmarks. }To evaluate our model's general visual understanding capabilities, we test it across several domains. These include general visual knowledge and hallucination mitigation (RealWorldQA~\citep{realworldqa}, Hallusion-Bench~\citep{guan2024hallusionbench}), perception and reasoning (BLINK~\citep{fu2024blink}, CharXiv-RQ~\citep{wang2024charxiv}), as well as document parsing and text-centric visual question answering (DocVQA~\citep{mathew2021docvqa}, OCRBench~\citep{liu2024ocrbench}, TextVQA~\citep{singh2019textvqa}). In Figure~\ref{fig:12-general_benchamrk}, we compare HY-Embodied-0.5 MoT-2B with two size-matched general VLMs, namely Qwen3-VL-2B-Thinking and InternVL 3.5-2B, on these benchmarks. The results show that while HY-Embodied-0.5 MoT-2B demonstrates strong performance in embodied and spatial understanding, it also achieves performance comparable to the size-matched general VLMs on general visual tasks.

\input{tab/general_benchmark_hymot2b}

\subsection{Results of HY-Embodied-0.5 MoE-A32B}

We evaluated our HY-Embodied-0.5 MoE A32B against several state-of-the-art visual agents, including Kimi K2.5~\citep{kimiteam2026kimik25visualagentic}, Seed 2.0~\citep{seed20}, Gemini 3.0 Pro~\citep{gemini3}, and Qwen 3.5 A17B~\citep{qwen35}, using the same benchmarking methodology described above. For Gemini 3.0 Pro and Seed 2.0, assessments were conducted via their official APIs under thinking mode. Results are summarized in Table~\ref{tab:hy32b_main_table}. Across 22 benchmarks, HY-Embodied-0.5 MoE A32B achieved first place in 7 tasks (32\%) and second place in 6 tasks (27\%), yielding an overall score of 67.0, outperforming Gemini 3.0 Pro by 3.4 points (vs. 63.6), Seed 2.0 by 0.8 points (vs. 66.2), Qwen 3.5 A17B by 0.9 point (vs. 66.1), and Kimi K2.5 by 5.9 points (vs. 61.1).

\input{tab/embodied_table_hya32b}

\subsection{Analysis}

In this subsection, we provide a detailed analysis of the HY-Embodied-0.5 model. We first present qualitative results on critical tasks involving visual perception and embodied environments. Then, leveraging our mix-chain architecture, we illustrate the chain-of-thought process to demonstrate the model's test-time scaling capabilities in long-chain mode. Finally, we validate our design choices through efficiency evaluations of the MoT architecture and attention visualizations of the visual latent tokens.

\begin{figure}[tb]
  \centering
  \includegraphics[width=\linewidth]{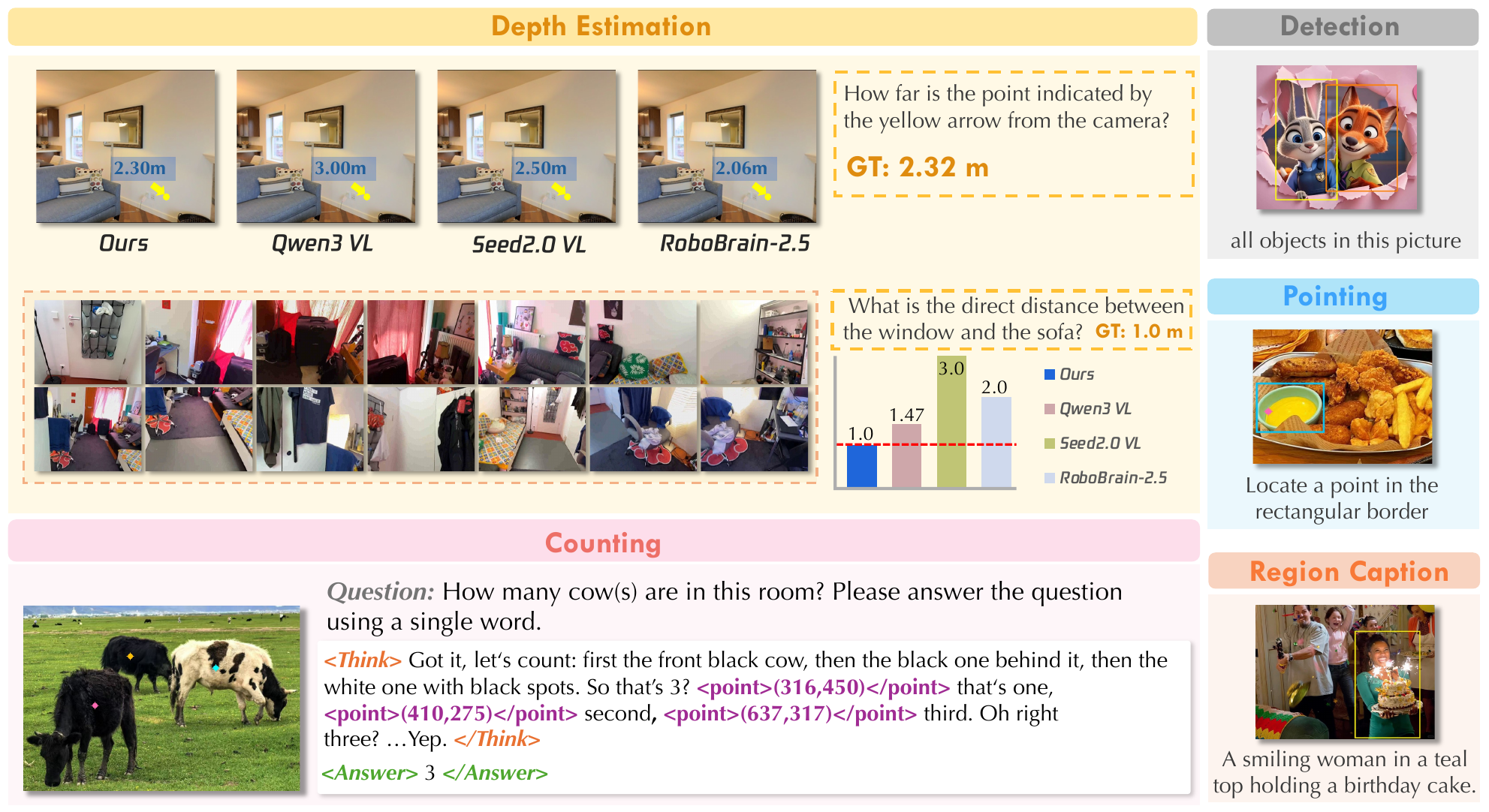}
  \caption{\textbf{Visualization Results on Visual Perception Tasks. }Empowered by our comprehensive visual perception training, HY-Embodied-0.5 MoT-2B demonstrates superior proficiency across foundational vision tasks, including depth estimation, object detection, and complex counting, outperforming competing embodied-specific and general VLMs.}
  \label{fig:7-perception}
\end{figure}

\paragrapha{Qualitative Results on Visual Perception Tasks.} Empowered by our large-scale, high-quality visual and embodied perception datasets, as well as comprehensive spatial recognition data, our model demonstrates robust proficiency across foundational visual tasks. As illustrated in Fig.~\ref{fig:7-perception}, in depth estimation scenarios—encompassing both the absolute distance from a specified point to the camera and the direct distance between objects across multiple views—our model yields predictions that are significantly closer to the Ground Truth (GT) compared to baseline models such as open-sourced model Qwen3 VL, proprietary model Seed2.0 VL, and embodied-specific model RoboBrain-2.5. Furthermore, in visual grounding tasks, the model exhibits high precision, delivering accurate results in bounding box detection, point-level localization, and region-level captioning. Notably, for complex counting tasks, our model effectively leverages a visual Chain-of-Thought (CoT) reasoning process. By sequentially identifying and assigning precise spatial coordinates to each target object during the reasoning phase, it logically deduces the accurate final answer. Collectively, these results underscore our model's exceptional capability in low-level visual perception, which inherently establishes a robust foundation for its superior performance in complex embodied environments.

\begin{figure}[tb]
  \centering
  \includegraphics[width=\linewidth]{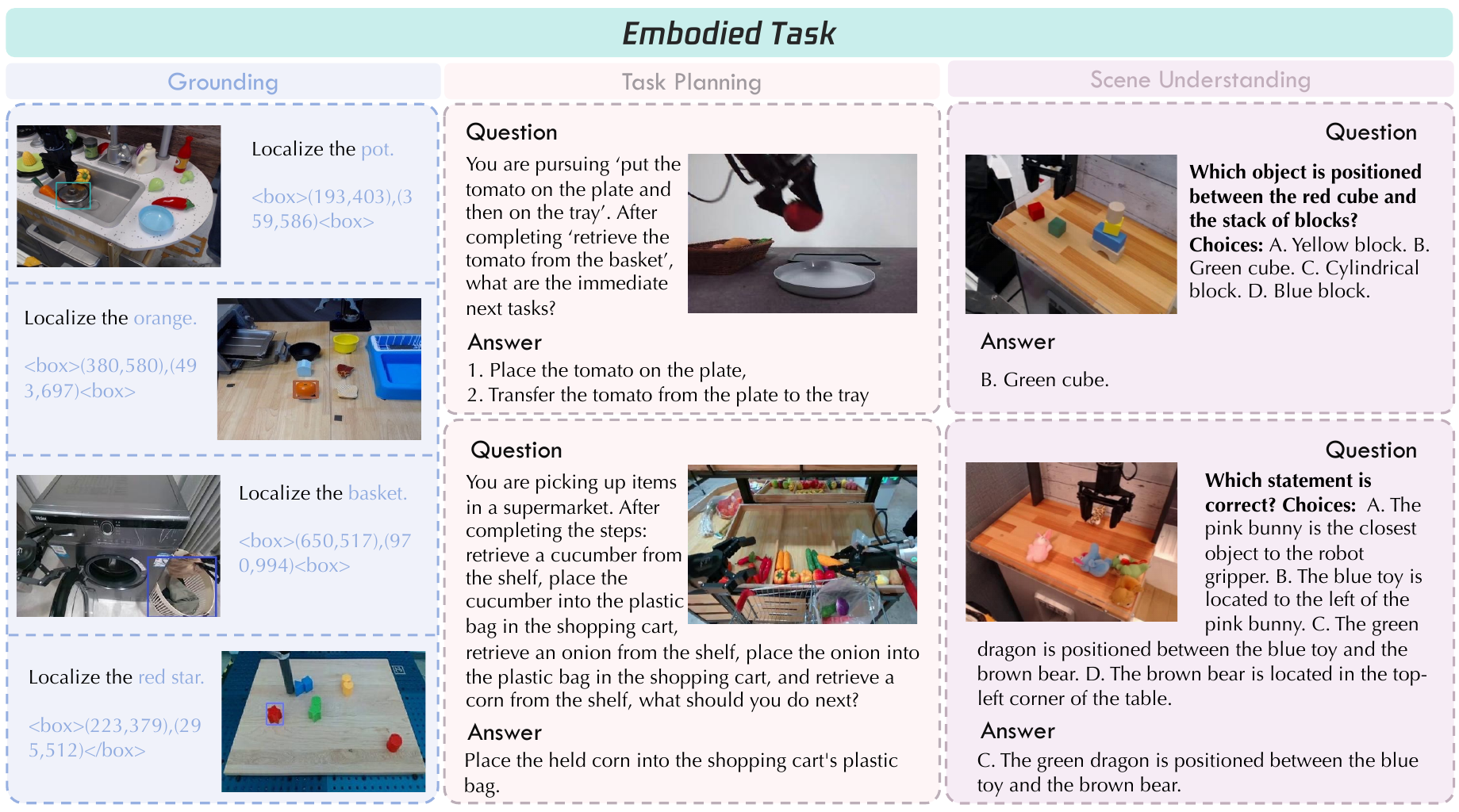}
  \caption{\textbf{Visualization Results on Embodied Tasks. }Our model demonstrates comprehensive proficiency across embodied tasks, including precise visual grounding, logical action planning, and scene understanding.}
  \label{fig:8-embodied}
\end{figure}

\paragrapha{Qualitative Results on Embodied Tasks.}Benefiting from the extensive and diverse embodied and spatial understanding data utilized during training, our model demonstrates comprehensive and highly accurate performance across multiple hierarchical levels of embodied tasks, specifically Embodied Perception (Grounding), Scene Understanding, and Task Planning. As illustrated in Fig.~\ref{fig:8-embodied}, in the Grounding task, the model exhibits precise localization capabilities, successfully outputting accurate bounding box coordinates for specific target objects (e.g., a pot, an orange, a basket, and a red star) amidst various cluttered robotic environments. For Scene Understanding, the model proves adept at parsing complex 3D spatial relationships. It accurately answers questions by correctly identifying objects based on their relative positions (such as locating a green cube between other blocks) and verifying intricate spatial statements among multiple items. Furthermore, in Task Planning scenarios, the model showcases strong sequential reasoning. Given a high-level objective and a history of completed steps, it accurately deduces the logical next actions—whether it involves determining the sequential placement of a tomato across different receptacles or inferring the next manipulation step in a multi-step supermarket picking task. More visualizations are provided in the Appendix.

\begin{figure}[tb]
  \centering
  \includegraphics[width=\linewidth]{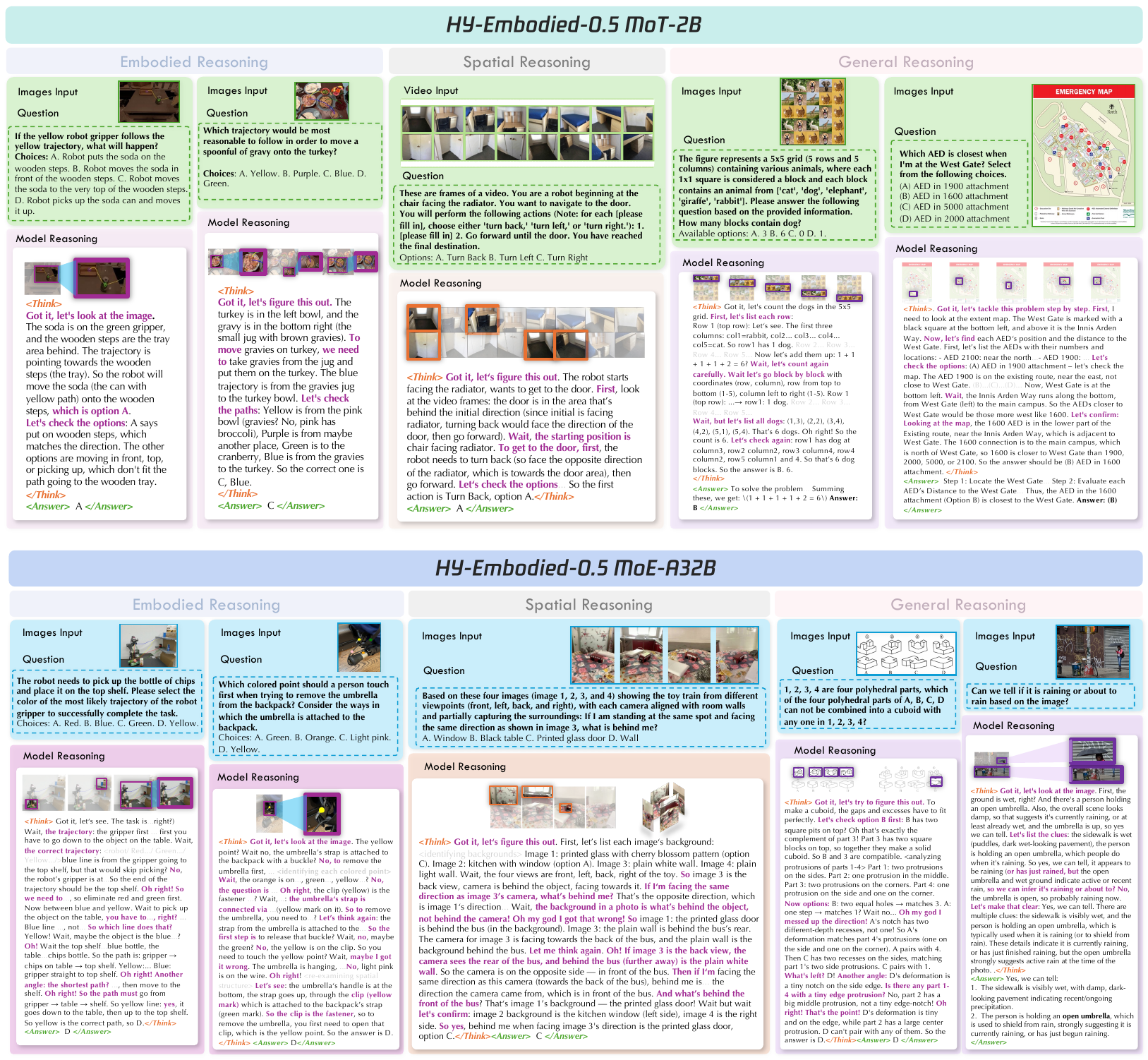}
  \caption{\textbf{Illustration of Chain-of-Thought Process.} Our HY-Embodied series demonstrates exceptional long-chain reasoning capabilities when tackling complex visual and embodied challenges. Regarding the specific thought process, rather than simply guessing outcomes, the models systematically analyze spatial relationships and affordances step-by-step, exhibiting advanced self-reflection and error-correction mechanisms within the thinking phase.}
  \label{fig:9-cot}
\end{figure}

\paragrapha{Illustration of CoT.} Empowered by our efficient, scientifically designed, multi-stage embodied post-training pipeline, our models exhibit exceptional long-chain reasoning capabilities. As illustrated in Fig.~\ref{fig:9-cot}, we showcase the profound ability of both the HY-Embodied-0.5 MoT-2B and A32B variants to resolve complex visual and embodied challenges through a robust Chain-of-Thought (CoT) process. Across Embodied Reasoning tasks, the models do not simply guess the final action; instead, they systematically analyze spatial relationships and affordances step-by-step—such as evaluating the correctness of different robot trajectories for manipulating objects or determining the precise interaction points for unbuckling a backpack. Notably, the $<think>$ process reveals advanced self-reflection and correction (e.g., explicitly pausing to reconsider structural details with phrases like "Wait, no..."). Furthermore, in Spatial and General Reasoning scenarios, the CoT mechanism enables the models to perform complex perspective-taking (inferring unseen environments from multi-view images), sequential navigation planning from video frames, and intricate 3D geometric deduction (matching polyhedral parts). These results demonstrate that our models engage in a transparent, logical, and highly reliable cognitive process when faced with complex, multi-step problems.

\begin{figure}[tb]
  \centering
  \includegraphics[width=\linewidth]{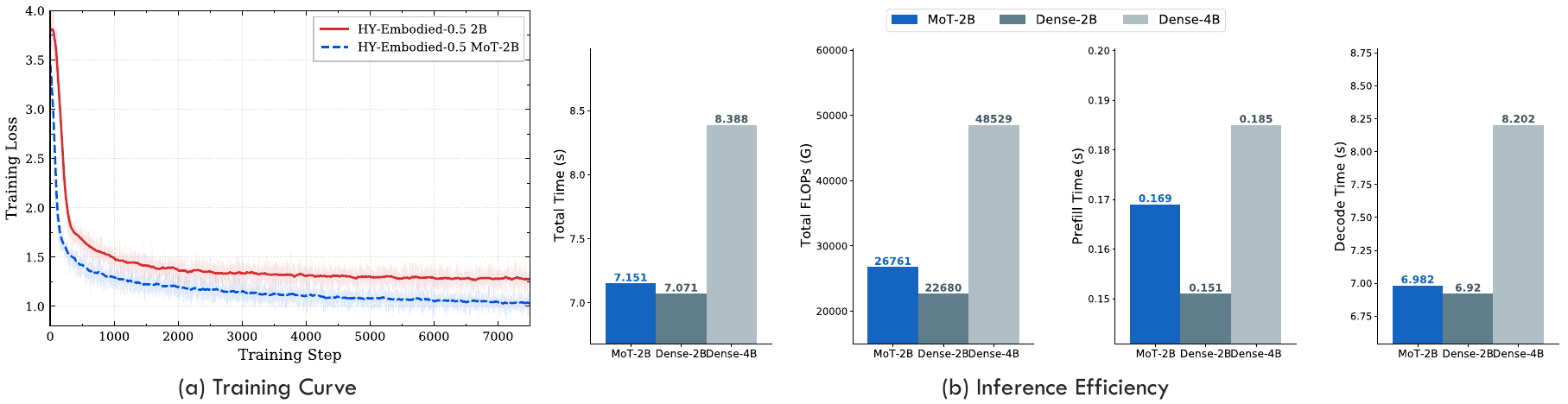}
  \caption{\textbf{MoT architecture enables faster convergence than the standard transformers (left), while delivering comparable inference speed (right). }(a) presents the training loss curves, and (b) details the inference efficiency by comparing the total inference time, theoretical total FLOPs, prefill time, and decode time across different models.}
  \label{fig:10-efficiency}
\end{figure}

\paragrapha{Efficiency for HY-Embodied-0.5 MoT.} The proposed Mixture-of-Tokens (MoT) architecture demonstrates highly desirable characteristics, achieving faster convergence and lower final loss during training, while introducing almost no additional overhead during inference. To ensure a fair comparison, both models are trained using identical training data, initialization methods, and hyperparameters. For the inference evaluation, we simulate practical real-world settings by fixing the input image tokens at 576 and the generated output tokens at 100. As illustrated in Figure~\ref{fig:10-efficiency}, we present the training loss curves with and without the MoT structure, alongside the actual total inference time and theoretical FLOPs. The training curves clearly show the efficiency of MoT, and during inference, the MoT architecture yields results closely approaching the Dense-2B baseline. Furthermore, we provide a detailed time breakdown for the prefill and decode stages. Because the decoding process dominates the total inference time in practical scenarios, the overall additional time overhead introduced by the MoT structure is negligible.

\begin{figure}[tb]
  \centering
  \includegraphics[width=0.95\linewidth]{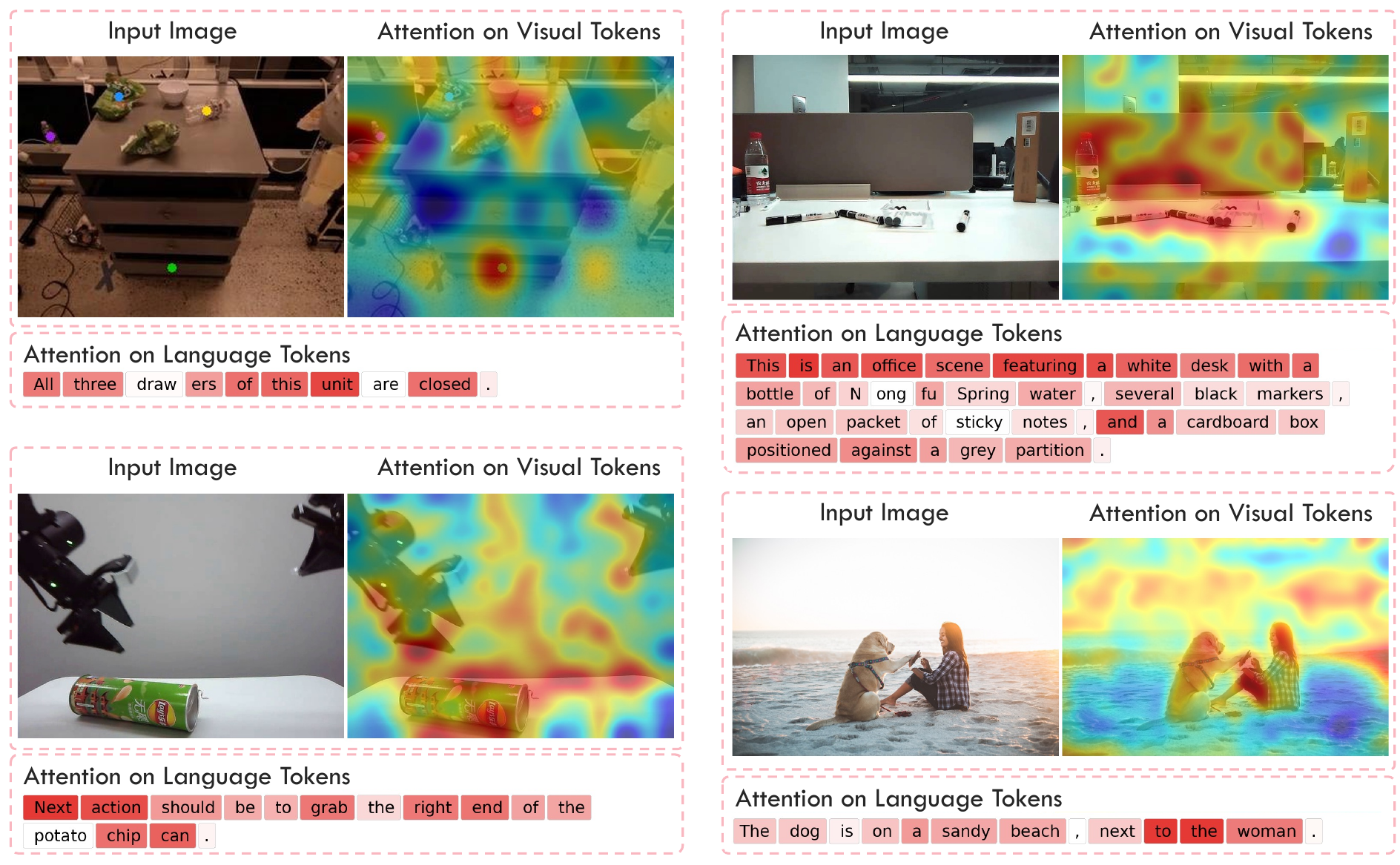}
  \caption{\textbf{Attention Visualizations for Visual Latent Tokens.} Visual attention accurately localizes salient objects and key spatial regions, while language attention concurrently focuses on the corresponding core semantic entities, states, and action instructions.}
  \label{fig:11-visual_latent_token}
\end{figure}

\paragrapha{Attention Visualizations for Visual Latent Tokens. }Visual latent tokens act as a connection between the visual full attention and the language causal attention. To demonstrate this, we provide the visualization for the attention map to visual tokens and the attention to text tokens in Fig.~\ref{fig:11-visual_latent_token}. We can observe from the figure that the visual attention maps precisely localize salient objects, highly specific object parts (such as the right end of the potato chip can or the handles of the drawers), and key spatial regions relevant to the scene context. Concurrently, the language attention weights are strongly concentrated on core semantic entities, state descriptions (e.g., "closed"), spatial relationships (e.g., "positioned against", "next to"), and action-oriented instructions (e.g., "grab"). This demonstrates that the visual latent tokens effectively bridge the modality gap by extracting fine-grained, semantically meaningful visual features and aligning them explicitly with the corresponding linguistic concepts. Consequently, our model exhibits a strong capacity to ground complex visual observations and embodied affordances into natural language, validating the effectiveness of our latent token design for cross-modal understanding and reasoning.

%% file: tab/embodied_table_hymot2b.tex
\begin{table}[t]
\caption{\textbf{Results for HY-Embodied-0.5 MoT-2B under 22 Embodied-Relevant Benchmarks.} We compare HY-Embodied-0.5 MoT-2B with existing state-of-the-art embodied foundation VLMs under 7B parameters across benchmarks for Embodied Understanding, Spatial Understanding, and Perception. We use \colorbox{bestcolor}{\hspace{1.2em}} and \colorbox{secondcolor}{\hspace{1.2em}} to denote the best and second-best results. Results for HY-Embodied-0.5 MoT-2B are reported in thinking mode, while for all other models, we report the better performance between non-thinking and thinking modes. }\vspace{5pt}
\renewcommand{\arraystretch}{1.5} 
\setlength{\tabcolsep}{3pt} 
\centering
\small
\begin{threeparttable}
\setlength{\tabcolsep}{1.8mm}{
\begin{adjustbox}{width=\textwidth}
\begin{tabular}{llc|cccc} \toprule
    \multirow[t]{3}{*}{\textbf{Capability}} & \multirow[t]{3}{*}{\textbf{Benchmark}} & \textbf{HY-Embodied} & \textbf{Qwen} & \textbf{Qwen} & \textbf{RoboBrain}  & \textbf{MiMo-Embodied} \\
    & & \textbf{0.5 MoT-2B} & \textbf{3-VL$^*$ 2B} & \textbf{3-VL$^*$ 4B} & \textbf{2.5 4B} & \textbf{7B} \\
    \midrule
    \multirow{2}{*}{\shortstack[l]{{\scriptsize Visual}\\{\scriptsize Preception}}} & CV-Bench & \cellcolor{bestcolor}\bf89.2 & 80.0 & 85.7 & 86.9 & \cellcolor{secondcolor}88.8 \\
    & DA-2K & \cellcolor{bestcolor}\bf92.3 & 69.5 & 76.5 & \cellcolor{secondcolor}79.4 & 72.2 \\
    \midrule
    \multirow{8}{*}{\shortstack[l]{{\scriptsize Embodied}\\{\scriptsize Understanding}}} 
    & ERQA & \cellcolor{bestcolor}\bf54.5 & 41.8 & \cellcolor{secondcolor}47.3  & 43.3 & 46.8 \\
    & EmbSpatial-Bench & \cellcolor{bestcolor}\bf82.8 & 75.9 & \cellcolor{secondcolor}80.7  & 73.8 & 76.2 \\
    & RoboBench-MCQ & \cellcolor{bestcolor}\bf49.2 & 36.9 & \cellcolor{secondcolor}45.8 & 44.4 & 43.6 \\
    & RoboBench-Planning & \cellcolor{secondcolor}54.2 & 36.2 & 36.4 & 39.2 & \cellcolor{bestcolor}\bf58.7 \\
    & RoboSpatial-Home & 55.7 & 45.3 & \cellcolor{bestcolor}\bf63.2  & \cellcolor{secondcolor}62.3 & 61.8 \\
    & ShareRobot-Aff. & \cellcolor{bestcolor}\bf 26.8 & 19.8 &  \cellcolor{secondcolor}25.5 & \cellcolor{secondcolor} 25.5 & 9.0 \\
    & ShareRobot-Traj. & \cellcolor{secondcolor}73.3 & 41.6 & 62.2  & \cellcolor{bestcolor}\bf81.4 & 50.6 \\
    & Ego-Plan2 & \cellcolor{secondcolor}45.5 & 35.5 & 38.8  & \cellcolor{bestcolor}\bf52.6 & 39.9 \\
    \midrule
    \multirow{12}{*}{\shortstack[l]{{\scriptsize Spatial}\\ {\scriptsize Understanding}}} & 3DSRBench & \cellcolor{bestcolor}\bf57.0 & 39.9 & 43.9 & \cellcolor{secondcolor}44.8 & 42.0 \\
    & All-Angles Bench & \cellcolor{bestcolor}\bf55.1 & 42.3 & 46.7 & 43.8 & \cellcolor{secondcolor}49.0 \\
    & MindCube & \cellcolor{bestcolor}\bf66.3 & 28.4 & 31.0 & 26.9 & \cellcolor{secondcolor}36.2 \\
    & MMSI-Bench & \cellcolor{bestcolor}\bf33.2 & 23.6 & 25.1 & 20.5 & \cellcolor{secondcolor}31.9 \\
    & RefSpatial-Bench & 45.8 & 28.9 & 45.3  & \cellcolor{bestcolor}\bf56.0 & \cellcolor{secondcolor}48.0 \\    
    & SAT & \cellcolor{secondcolor}76.7 & 45.3 & 56.7  & 51.3 & \cellcolor{bestcolor}\bf78.7 \\
    & SIBench-mini & \cellcolor{bestcolor}\bf58.2 & 42.0 & 50.9 & 47.3 & \cellcolor{secondcolor}53.1 \\
    & SITE-Bench-Image & \cellcolor{bestcolor}\bf62.7 & 52.3 & \cellcolor{secondcolor}61.0 & 57.9 & 49.9 \\
    & SITE-Bench-Video & \cellcolor{bestcolor}\bf63.5 & 52.2 & 58.0  & 54.8 & \cellcolor{secondcolor}58.9 \\
    & ViewSpatial & \cellcolor{bestcolor}\bf53.1 & 37.2 & \cellcolor{secondcolor}41.6  & 36.6 & 36.1 \\
    & VSIBench & \cellcolor{bestcolor}\bf60.5 & 48.0 & \cellcolor{secondcolor}55.2 & 41.7 & 48.5 \\
    & Where2Place & \cellcolor{bestcolor}\bf68.0 & 45.0 & 59.0  & \cellcolor{secondcolor}65.0 & 63.6 \\
    \bottomrule
\end{tabular}
\end{adjustbox}}
\end{threeparttable}
\label{tab:hymot2b_main_table}
\end{table}
\footnotetext{$^*$ We observe that small models from Qwen3.5 series produce repetitive thinking patterns in some benchmarks and leads to a lower overall results, so we compare Qwen3-VL models in our evaluations.}

%% file: tab/general_benchmark_hymot2b.tex
\begin{figure}[tb]
  \centering
  \includegraphics[width=\linewidth]{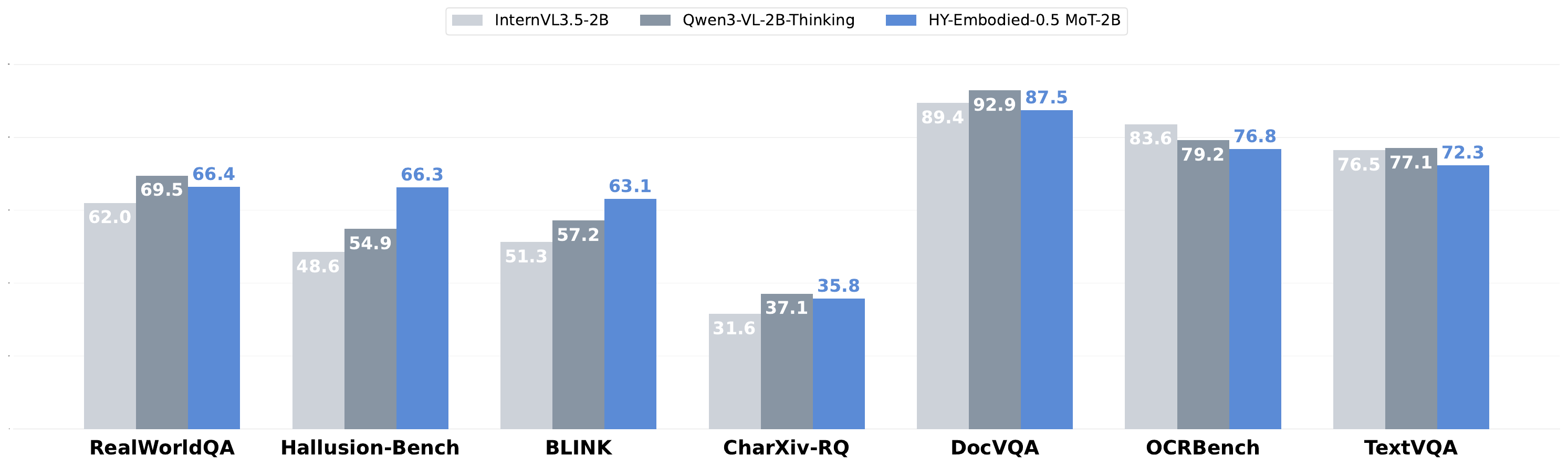}
  \caption{\textbf{Performance on General Understanding Benchmark. }Comparison of HY-Embodied-0.5 MoT-2B with size-matched general VLMs. The results demonstrate that while our model is specifically optimized for spatial and embodied reasoning, it successfully maintains comparable and highly competitive performance across diverse general visual understanding tasks.}
  \label{fig:12-general_benchamrk}
\end{figure}

%% file: tab/embodied_table_hya32b.tex
\begin{table}[t]
\caption{\textbf{Results for HY-Embodied-0.5 MoE-A32B Compared with Existing Frontier VLMs. }We evaluate our model against state-of-the-art agents across 22 benchmarks under visual perception, embodied, and spatial understanding. We use \colorbox{bestcolor}{\hspace{1.2em}} and \colorbox{secondcolor}{\hspace{1.2em}} to denote the best and second-best results.}\vspace{5pt}
\renewcommand{\arraystretch}{1.5} 
\setlength{\tabcolsep}{3pt} 
\centering
\small
\begin{threeparttable}
\setlength{\tabcolsep}{1.8mm}{
\begin{adjustbox}{width=\textwidth}
\begin{tabular}{llc|cccc} \toprule
    \multirow[t]{3}{*}{\textbf{Capability}} & \multirow[t]{3}{*}{\textbf{Benchmark}} & \textbf{HY-Embodied} &     \ \ \ \ \ \textbf{Kimi}  \ \ \ \ \ & \ \ \ \ \ \textbf{Seed}  \ \ \ \ \ & \ \ \ \ \ \textbf{Qwen}  \ \ \ \ \ & \ \ \ \ \ \textbf{Gemini}\ \ \ \ \ \\
    & & \textbf{0.5 MoE A32B} & \textbf{K2.5} & \textbf{2.0} & \textbf{3.5 A17B} & \textbf{3.0 Pro} \\
    \midrule
    \multirow{2}{*}{\shortstack[l]{{\scriptsize Visual}\\{\scriptsize Preception}}} & CV-Bench & \cellcolor{secondcolor} 88.8 & \cellcolor{bestcolor}\bf 89.0 & 88.5$^*$ & 88.6 & 85.4$^*$ \\
    & DA-2K & \cellcolor{secondcolor} 90.2 & 83.4 & \cellcolor{bestcolor}\bf 92.3$^*$ & 83.3 & 83.6$^*$ \\
    \midrule
    \multirow{8}{*}{\shortstack[l]{{\scriptsize Embodied}\\{\scriptsize Understanding}}} 
    & ERQA & \cellcolor{secondcolor} 62.3 & 59.8 & 61.8$^*$ & 61.0 & \cellcolor{bestcolor}\bf 65.0$^*$ \\
    & EmbSpatial-Bench & \cellcolor{bestcolor}\bf 84.1 & 81.5 & 81.0$^*$ & \cellcolor{secondcolor} 83.8 & 83.6$^*$ \\
    & RoboBench-MCQ & 62.8 & 59.0 & \cellcolor{secondcolor} 66.5$^*$ & 63.8 & \cellcolor{bestcolor}\bf 69.2$^*$ \\
    & RoboBench-Planning & 59.3 & \cellcolor{secondcolor} 60.0 & \cellcolor{bestcolor}\bf 60.1$^*$ & 56.7 & \cellcolor{secondcolor} 60.0$^*$ \\
    & RoboSpatial-Home & \cellcolor{bestcolor}\bf 76.6 & 66.0 & 71.7$^*$ & \cellcolor{secondcolor} 74.9 & 57.1$^*$ \\
    & ShareRobot-Aff. & \cellcolor{secondcolor} 28.6 & 21.5 & 27.5$^*$ & \cellcolor{bestcolor}\bf 29.3 & 24.8$^*$ \\
    & ShareRobot-Traj. & \cellcolor{bestcolor}\bf 76.9 & 68.5 & 71.8$^*$ & \cellcolor{secondcolor} 73.8 & 68.7$^*$ \\
    & Ego-Plan2 & 51.4 & 47.4 & \cellcolor{secondcolor} 56.6$^*$ & 55.3 & \cellcolor{bestcolor}\bf 60.0$^*$ \\
    \midrule
    \multirow{12}{*}{\shortstack[l]{{\scriptsize Spatial}\\ {\scriptsize Understanding}}} & 3DSRBench & 56.6 & 55.9 & \cellcolor{secondcolor} 58.2$^*$ & 56.6 & \cellcolor{bestcolor}\bf 58.3$^*$ \\
    & All-Angles Bench & 71.8 & 64.8 & 69.3$^*$ & \cellcolor{secondcolor} 72.1 & \cellcolor{bestcolor}\bf 73.4$^*$ \\
    & MindCube & \cellcolor{bestcolor}\bf 69.2 & 57.8 & 55.2$^*$ & 59.0 & \cellcolor{secondcolor} 66.0$^*$ \\
    & MMSI-Bench & 39.2 & 36.5 & \cellcolor{secondcolor} 47.6$^*$ & 43.8 & \cellcolor{bestcolor}\bf 48.0$^*$ \\
    & RefSpatial-Bench & 57.2 & 43.3 & \cellcolor{bestcolor}\bf 72.2$^*$ & \cellcolor{secondcolor} 61.0 & 33.2$^*$ \\    
    & SAT & \cellcolor{secondcolor}87.3 & 79.3 &  86.2$^*$ & 86.0 & \cellcolor{bestcolor}\bf 88.0$^*$ \\
    & SIBench-mini & \cellcolor{secondcolor} 67.3 & 63.0 & 65.9$^*$ & 66.3 & \cellcolor{bestcolor}\bf 68.0$^*$ \\
    & SITE-Bench-Image & 74.7 & 73.8 & \cellcolor{secondcolor} 75.6$^*$ & \cellcolor{bestcolor}\bf 77.1 & 75.4$^*$ \\
    & SITE-Bench-Video & \cellcolor{bestcolor}\bf 72.5 & 71.5 & 68.9$^*$ & \cellcolor{secondcolor} 72.3 & 69.8$^*$ \\
    & ViewSpatial & \cellcolor{bestcolor}\bf 59.8 & 45.2 & \cellcolor{secondcolor} 56.4$^*$ & 52.2 & 50.8$^*$ \\
    & VSIBench & \cellcolor{bestcolor}\bf 68.3 & 54.2 & 51.0$^*$ & \cellcolor{secondcolor} 61.1 & 57.9$^*$ \\
    & Where2Place & 70.0 & 64.0 & \cellcolor{secondcolor} 73.0$^*$ & \cellcolor{bestcolor}\bf 76.0 & 52.0$^*$ \\
    \bottomrule
\end{tabular}
\end{adjustbox}}
\begin{tablenotes}
    \footnotesize
    \item[*] Results self-collected via API in March 2026.
\end{tablenotes}
\end{threeparttable}
\label{tab:hy32b_main_table}
\end{table}

%% file: sec/7-VLA.tex
\section{Robot Control Results}
\label{vla}

Building upon the MoT architecture of HY-Embodied-0.5-MoT-2B base model, we extend the Action Expert module following the structural design of $\pi0$/$\pi0.5$, resulting in the Vision-Language-Action (VLA) model for robot control experiments in real-world scenarios. 

To better unlock the potential of the VLA on real-robot tasks, we first fine-tune the network using 5K hours of UMI data. Since all training data in this stage originates from UMI, the network has not been exposed to any specific robot embodiment during this process. We use a per-GPU batch size of 32 across 32 GPUs, for a total of 200K iterations.

We then perform supervised fine-tuning (SFT) with real-robot data on the following three tasks and conduct deployment evaluations. Depending on the difficulty of each task, varying amounts of demonstration data are collected, ranging from 300 to 700 episodes. As baselines, $\pi0$ and $\pi0.5$ undergo SFT under identical conditions using the same real-robot data, with both the data volume and training iterations.

\begin{figure}[t]
    \centering
    
    \begin{subfigure}[b]{0.45\textwidth}
        \centering
        \includegraphics[width=\textwidth]{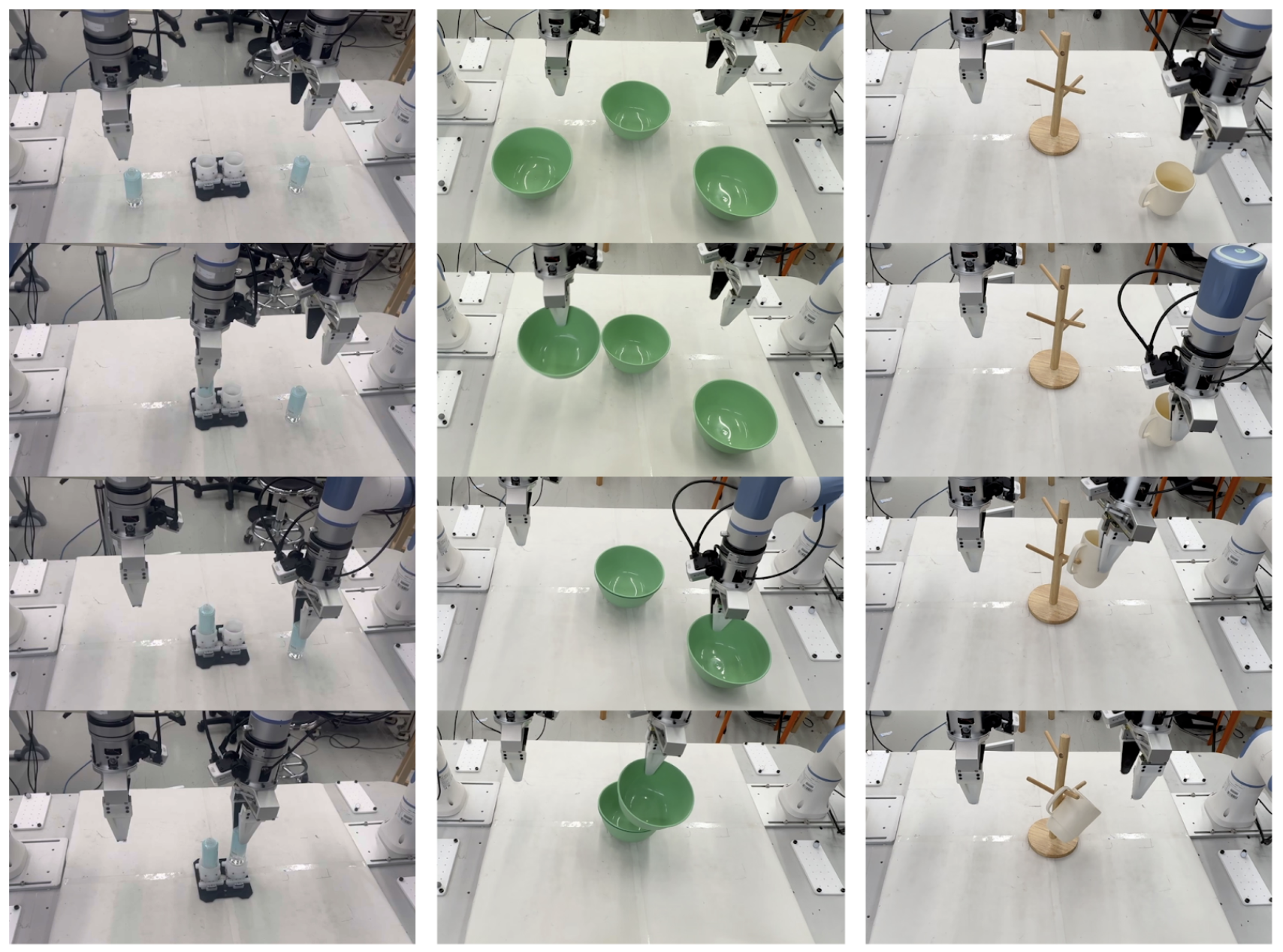}
        \caption{} 
        \label{fig:robot_setup}
    \end{subfigure}    
    \begin{subfigure}[b]{0.53\textwidth}
        \centering
        \includegraphics[width=\textwidth]{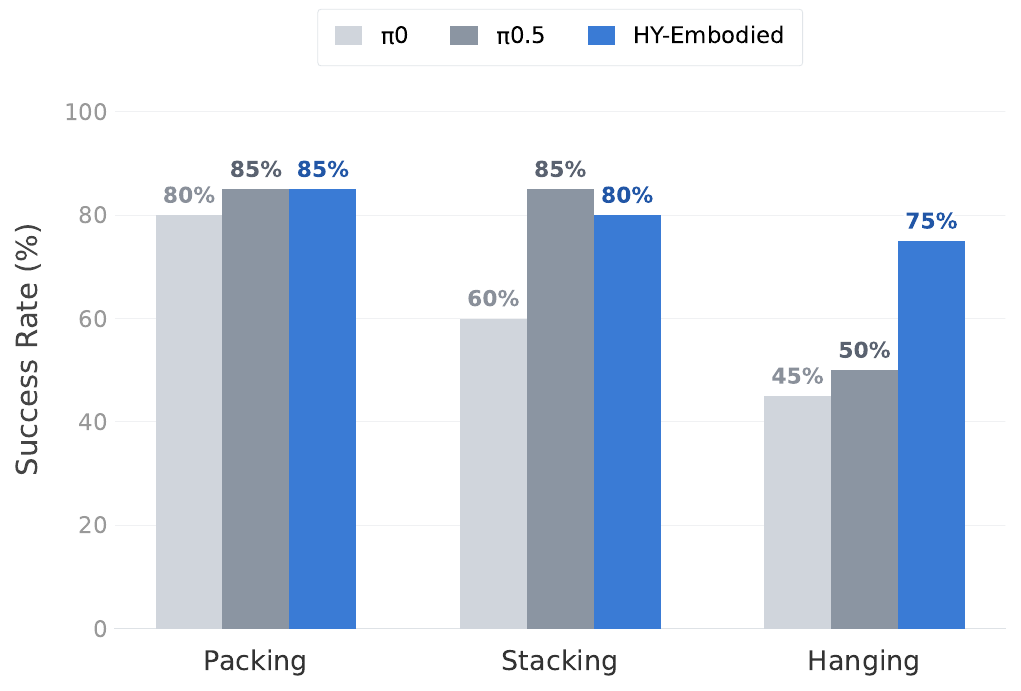}
        \caption{} 
        \label{fig:success_rate}
    \end{subfigure}
    
    \caption{\textbf{Robot Experimental Setup and Success Rates for the Evaluated Tasks.} 
    \textbf{(a)} Real-world setup of three representative tasks. Our platform employs a dual-arm Xtrainer equipped with head-mounted and wrist-mounted cameras. The benchmark includes: (1) Precision Plug-in Packing, (2) Tableware Stacking, and (3) Mug Hanging. Success rates are evaluated over 20 trials per task. 
    \textbf{(b)} Object poses are randomly initialized at the start of each trial. We conduct 20 real-robot trials per task per model. The success rates (\%) are summarized above.}
    \label{fig:robot_experiment}
\end{figure}

The robot experimental setup and results are shown in Figure~\ref{fig:robot_experiment}. As illustrated, our HY-Embodied-0.5 VLA model demonstrates robust and highly competitive performance across all three evaluated real-world tasks compared to the $\pi0$ and $\pi0.5$ baselines. For the Precision Plug-in Packing task, HY-Embodied-0.5 achieves a success rate of 85\%, matching the performance of $\pi0.5$ and surpassing $\pi0$ (80\%). In the Tableware Stacking task, our model attains an 80\% success rate, which is a substantial improvement over the 60\% achieved by $\pi0$ and remains competitive with the 85\% success rate of $\pi0.5$. Most notably, in the Mug Hanging task---which appears to be the most challenging given the baseline performances---HY-Embodied-0.5 demonstrates superior control capabilities, achieving a success rate of 75\%. This represents a significant margin of improvement over both $\pi0$ (45\%) and $\pi0.5$ (50\%). These compelling results suggest that the initial fine-tuning on the extensive 5K-hour UMI dataset, combined with the underlying MoT architecture, successfully equips the model with rich, generalizable representations that effectively transfer to complex, embodiment-specific manipulation tasks following supervised fine-tuning.

%% file: sec/8-Conclusion.tex
\section{Conclusion}

In this report, we propose HY-Embodied-0.5, a strong foundation vision-language model designed for real-world embodied tasks. HY-Embodied-0.5 represents a vital step forward in bridging the divide between general VLMs and the dynamic demands of real-world agents. By pioneering a modality-adaptive Mixture-of-Transformers (MoT) architecture alongside visual latent tokens, the model achieves the fine-grained spatial and visual perception required for physical grounding. Furthermore, its embodied post-training pipeline successfully compresses deep, complex reasoning capabilities into a highly efficient 2B parameter variant tailored for edge deployment. Ultimately, the suite's state-of-the-art performance across 22 demanding benchmarks and its robust execution in real-world robotic manipulation tasks demonstrate that HY-Embodied-0.5 effectively translates expansive digital intelligence into tangible, physical-world competence. We aim to further explore and bridge the gap between language and action models, ultimately training a real-world brain that is more conducive to complex real-world applications.

%% file: sec/9-Appendix.tex
\section{Visualizations}

\noindent
\textbf{Trajectory Prediction under End-Effector Control.}

\begin{center}
    \includegraphics[width=0.92\textwidth]{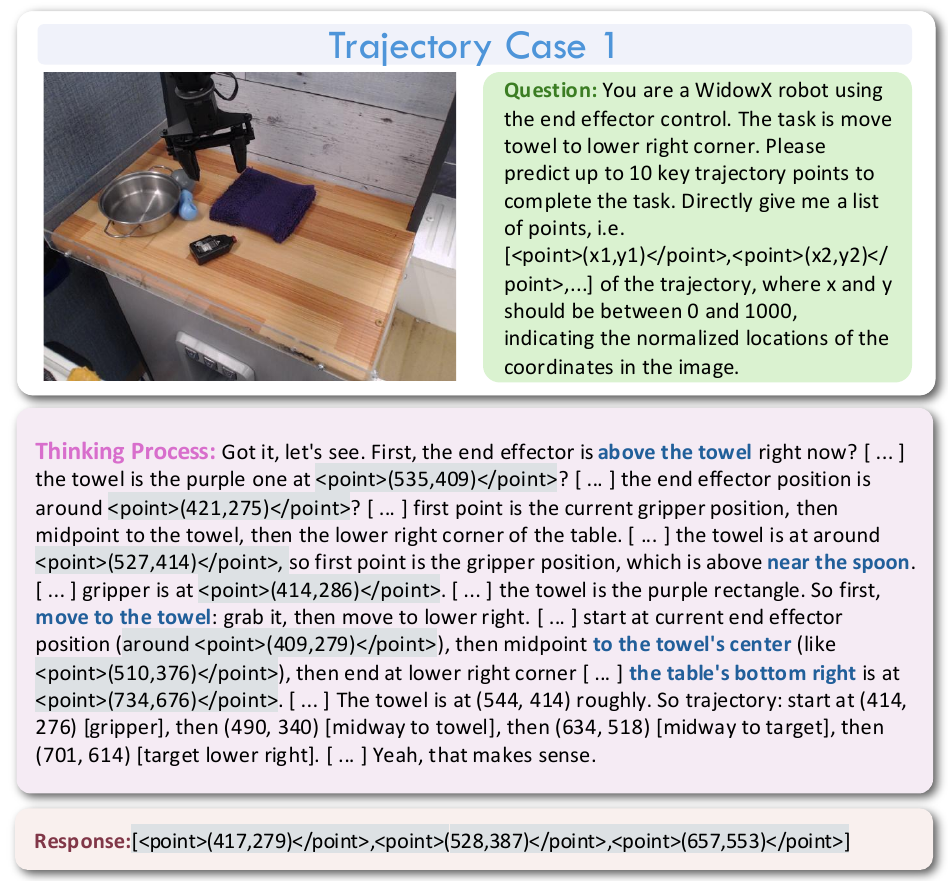}

    \vspace{0.5em}
    \small Fig.B1: \textbf{Example of end-effector trajectory prediction.}
\end{center}

\clearpage
\noindent
\textbf{Trajectory Prediction under Joint Control.}

\begin{center}
    \includegraphics[width=0.92\textwidth]{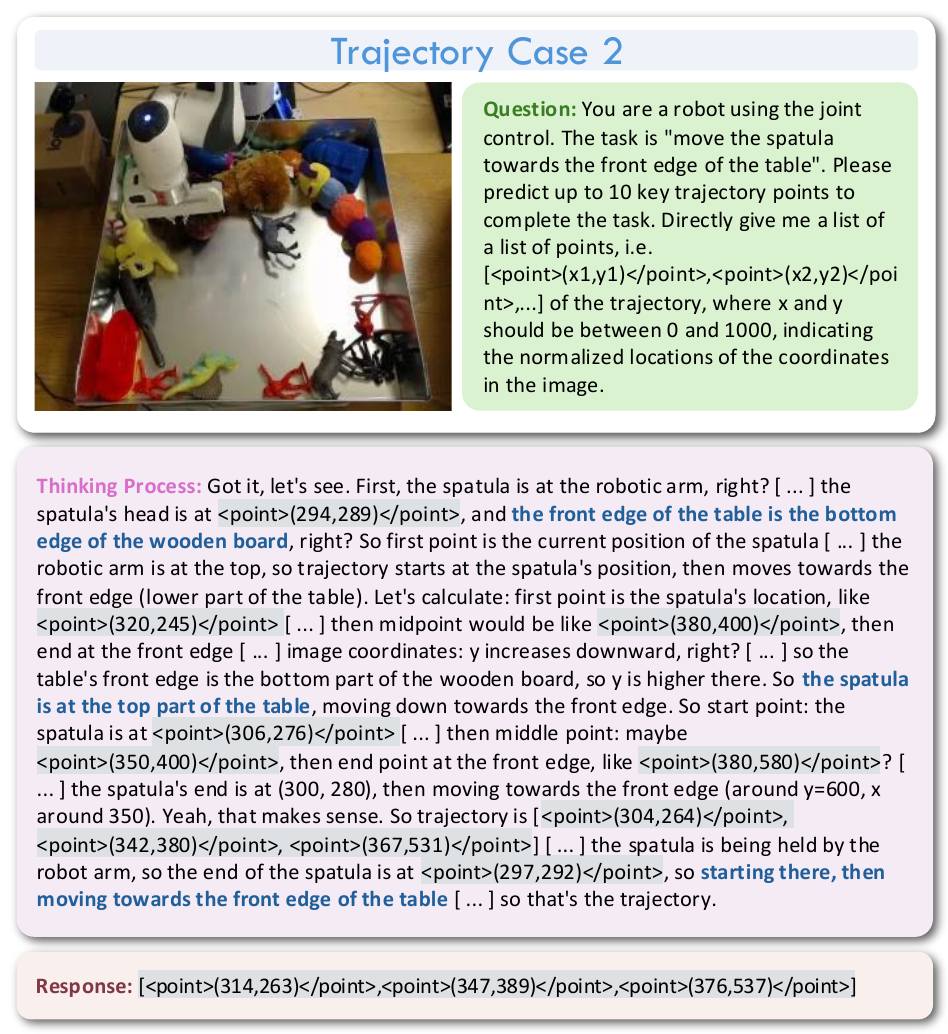}

    \vspace{0.5em}
    \small Fig.B2: \textbf{Example of joint-control trajectory prediction.} 
\end{center}

\clearpage
\noindent
\textbf{2D Bounding Box Grounding.}

\begin{center}
    \includegraphics[width=0.92\textwidth]{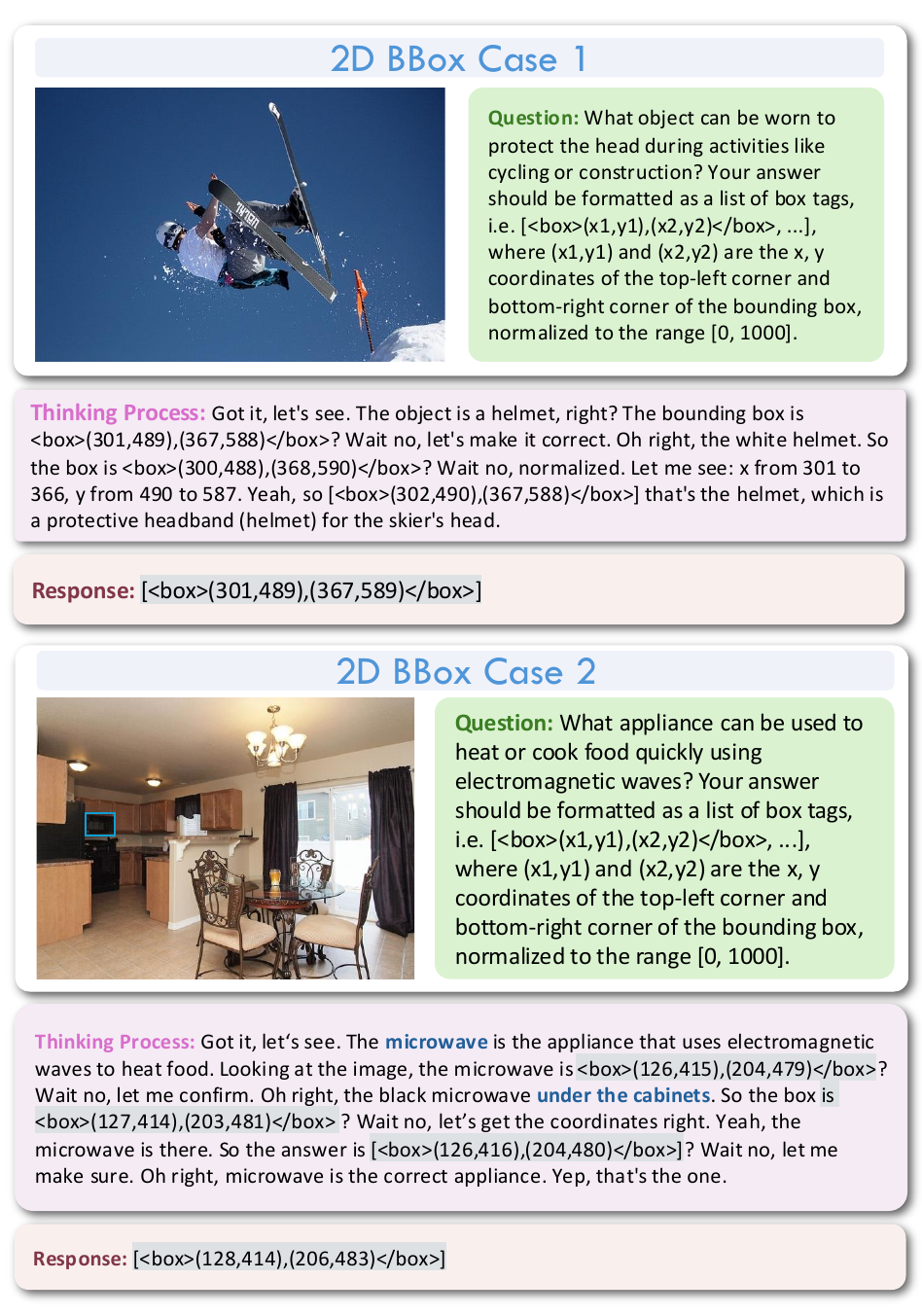}

    \vspace{0.5em}
    \small Fig.B3: \textbf{ Examples of 2D bounding box grounding.}
\end{center}

\clearpage
\noindent
\textbf{Point-Based Localization.}

\begin{center}
    \includegraphics[width=0.92\textwidth]{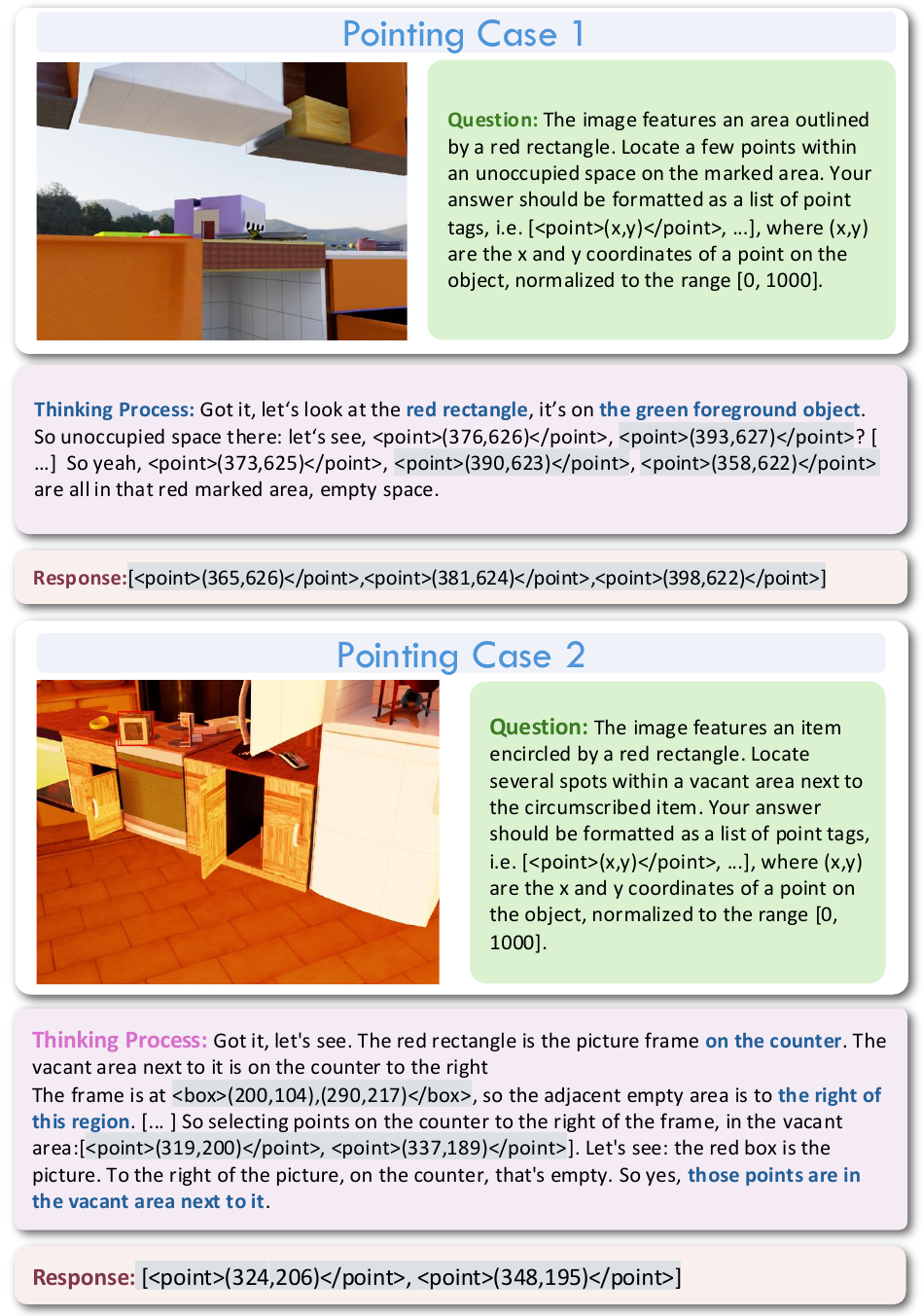}

    \vspace{0.5em}
    \small Fig.B4: \textbf{Examples of point-based localization.}  
\end{center}

\clearpage
\noindent
\textbf{Metric Reasoning: Distance Estimation.}

\begin{center}
    \includegraphics[width=0.92\textwidth]{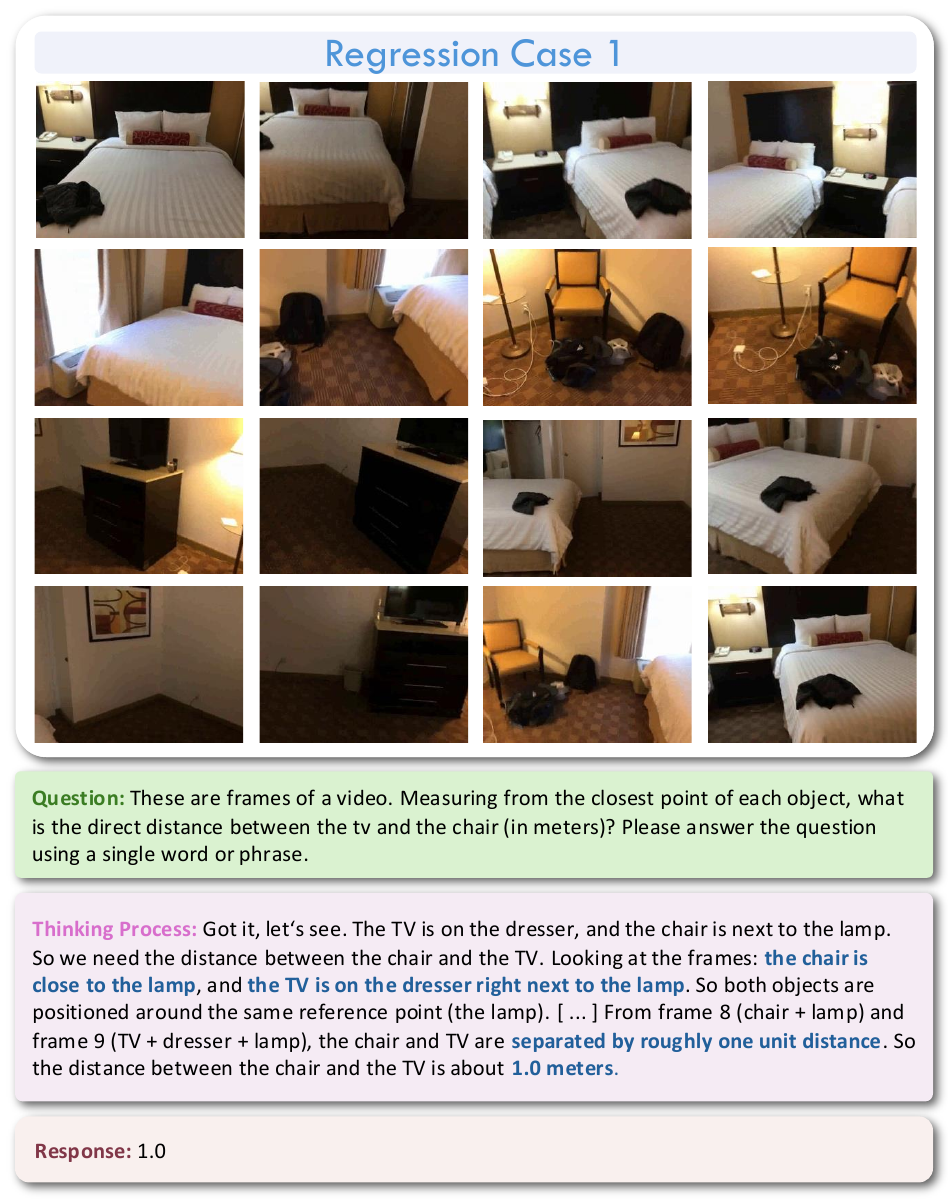}

    \vspace{0.5em}
    \small Fig.B5: \textbf{Example of distance estimation from multi-frame visual input.} 
\end{center}

\clearpage
\noindent
\textbf{Metric Reasoning: Area Estimation.}

\begin{center}
    \includegraphics[width=0.92\textwidth]{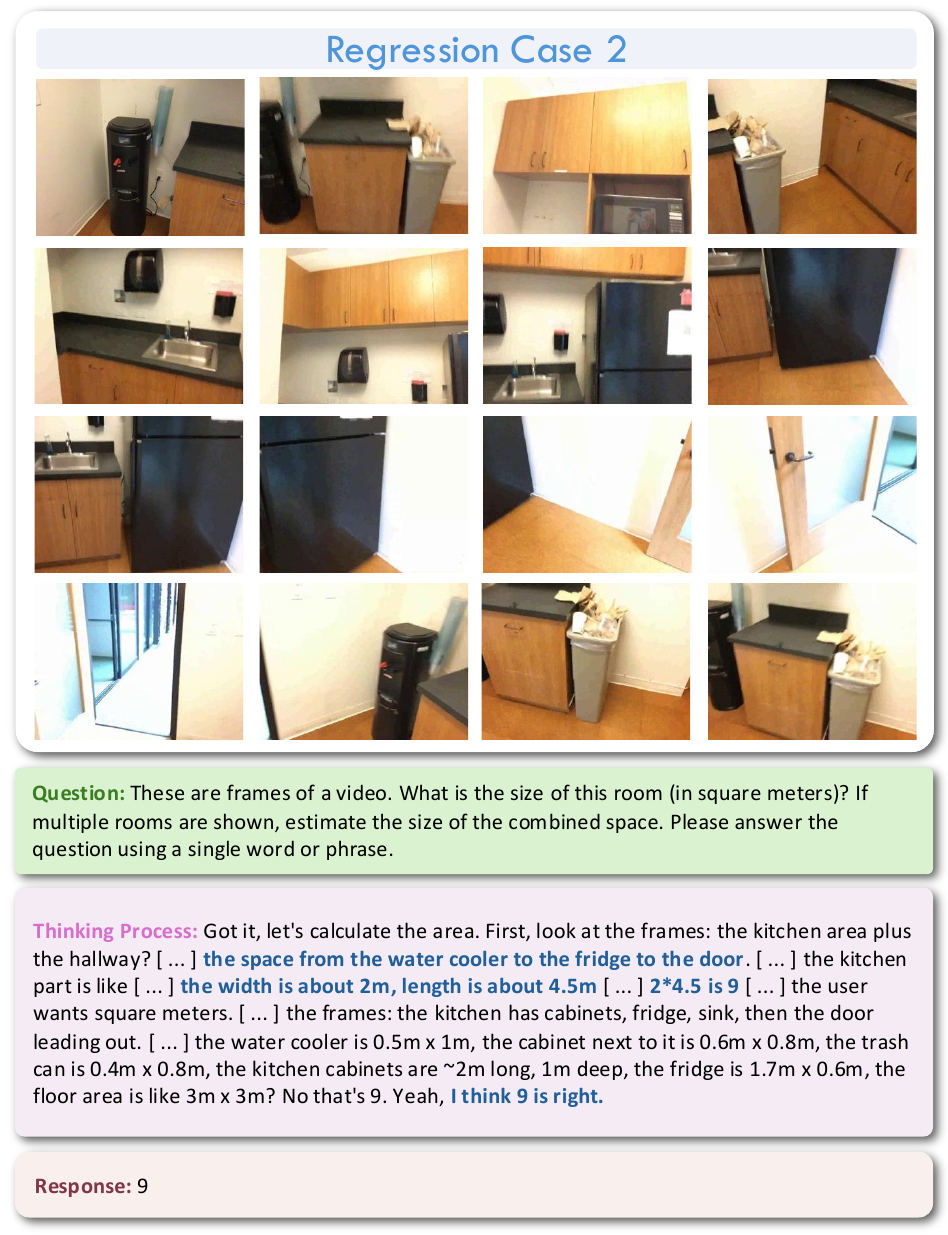}

    \vspace{0.5em}
    \small Fig.B6: \textbf{ Example of room-area estimation.}
\end{center}

\clearpage
\noindent
\textbf{Multi-View Spatial Reasoning.}

\begin{center}
    \includegraphics[width=0.92\textwidth]{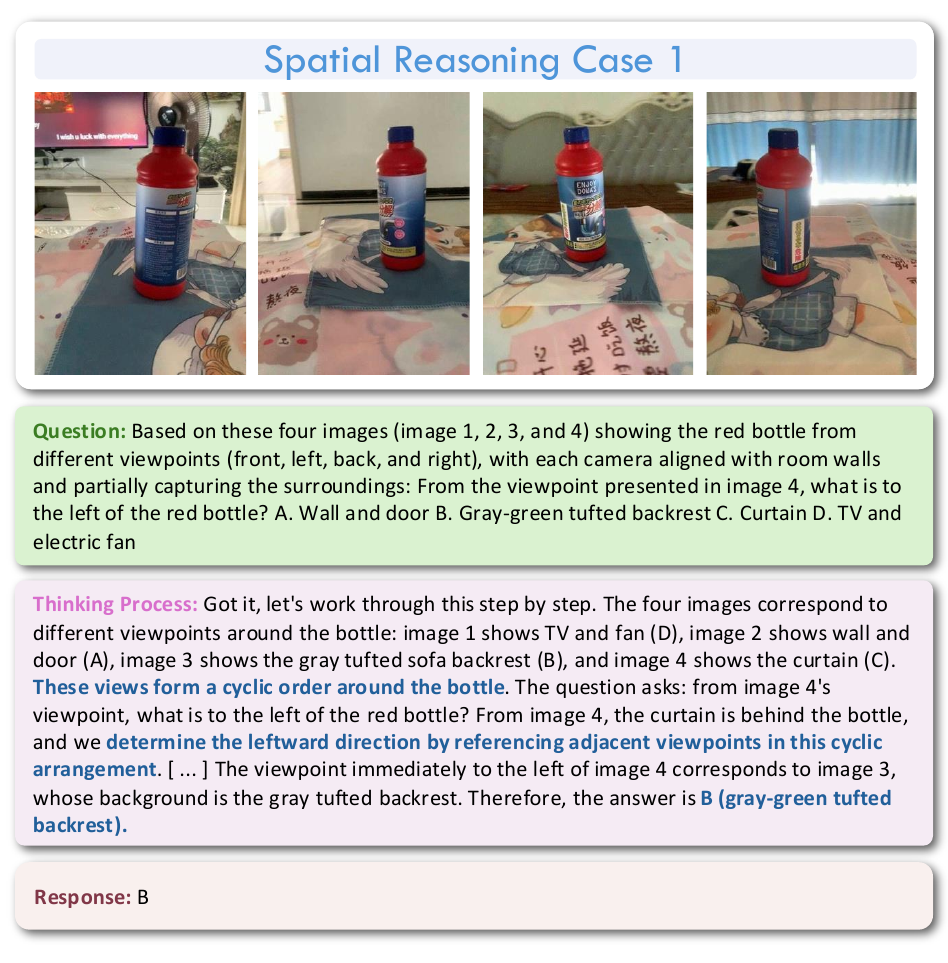}

    \vspace{0.5em}
    \small Fig.B7: \textbf{Example of multi-view spatial reasoning.}  
\end{center}

\clearpage
\noindent
\textbf{Orientation-Aware Spatial Reasoning.}

\begin{center}
    \includegraphics[width=0.92\textwidth]{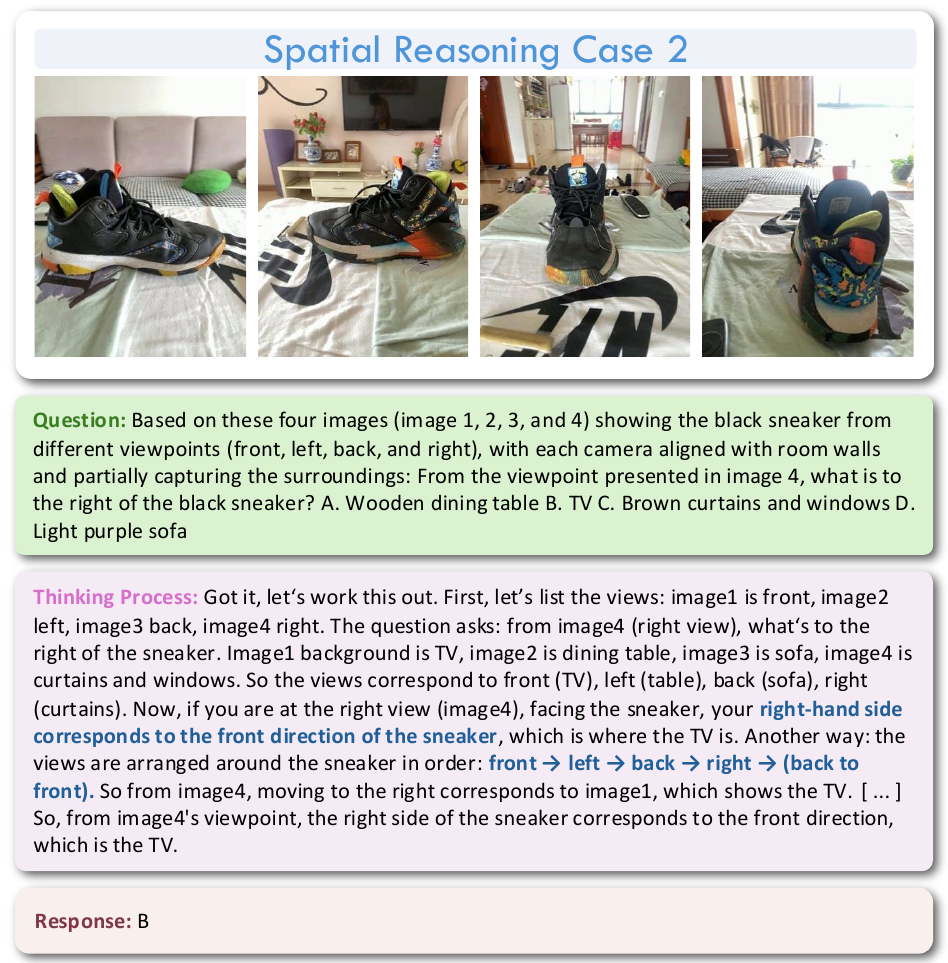}

    \vspace{0.5em}
    \small Fig.B8: \textbf{Example of orientation-aware spatial reasoning. } 
\end{center}

\clearpage
\noindent
\textbf{Embodied Perception.}

\begin{center}
    \includegraphics[width=0.92\textwidth]{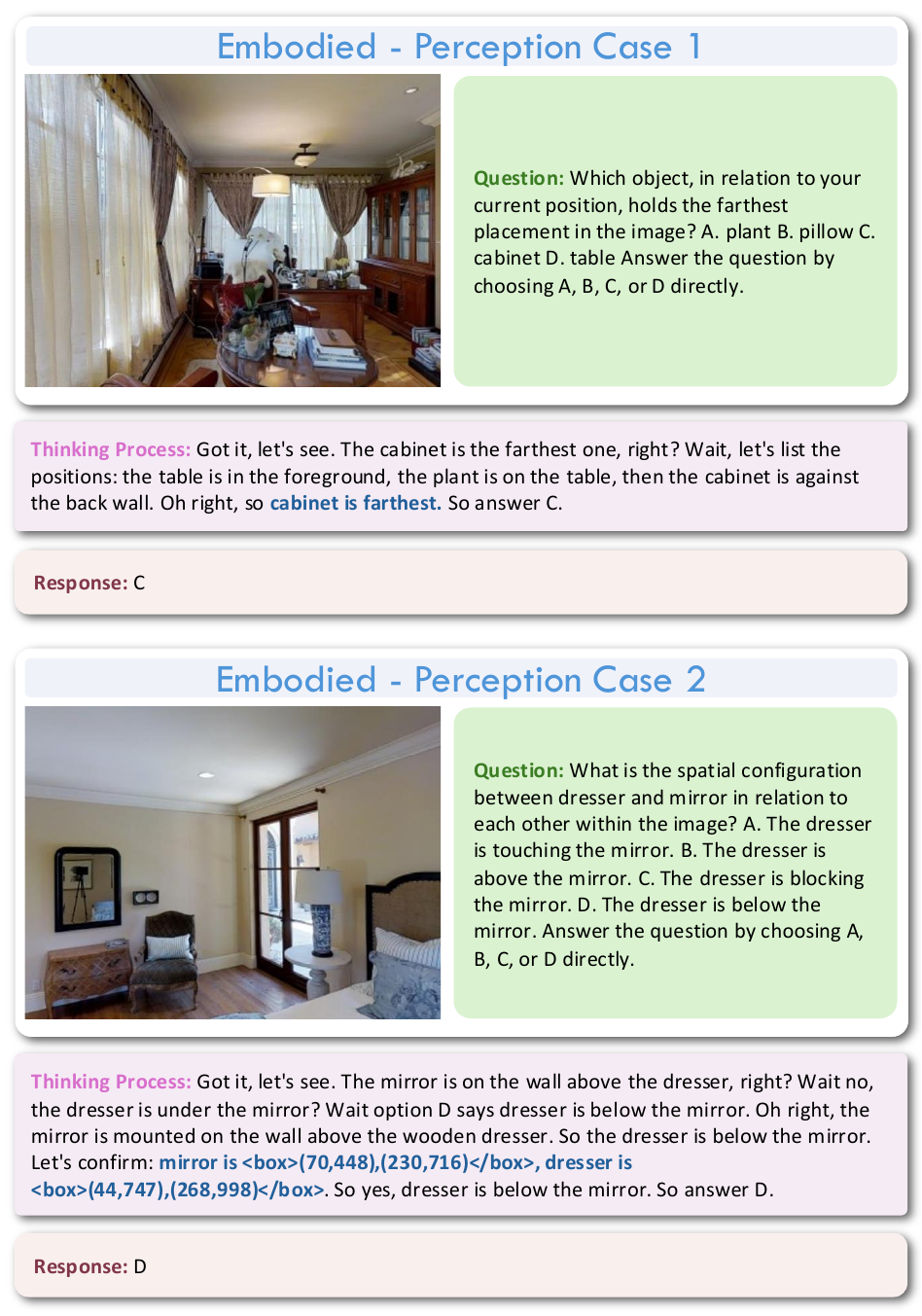}

    \vspace{0.5em}
    \small Fig.B9: \textbf{Examples of embodied perception.} 
\end{center}

\clearpage
\noindent
\textbf{Affordance-Aware Localization.}

\begin{center}
    \includegraphics[width=0.92\textwidth]{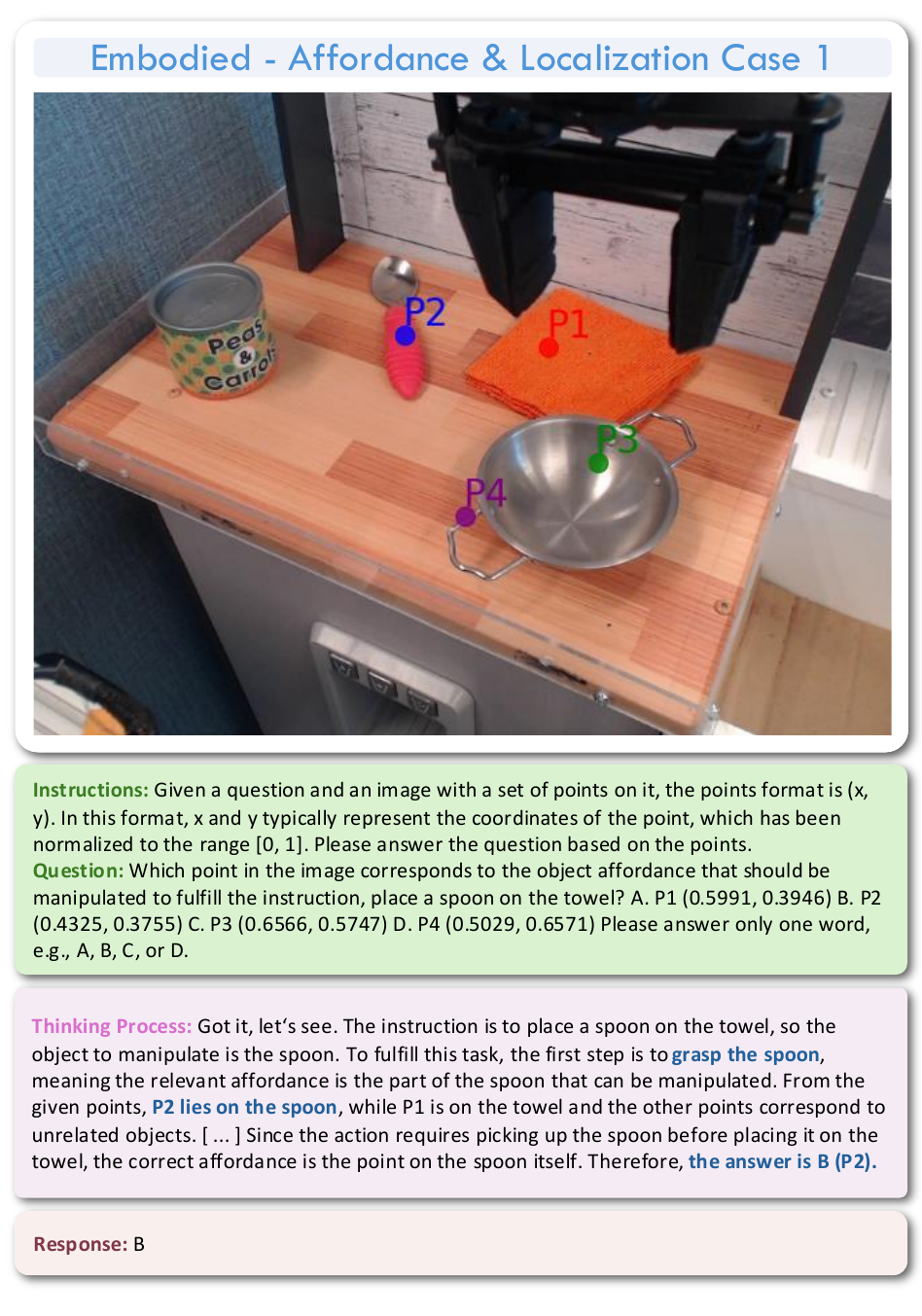}

    \vspace{0.5em}
    \small Fig.B10: \textbf{Example of affordance-aware localization.} 
\end{center}

\clearpage
\noindent
\textbf{Navigation-Oriented Localization.}

\begin{center}
    \includegraphics[width=0.92\textwidth]{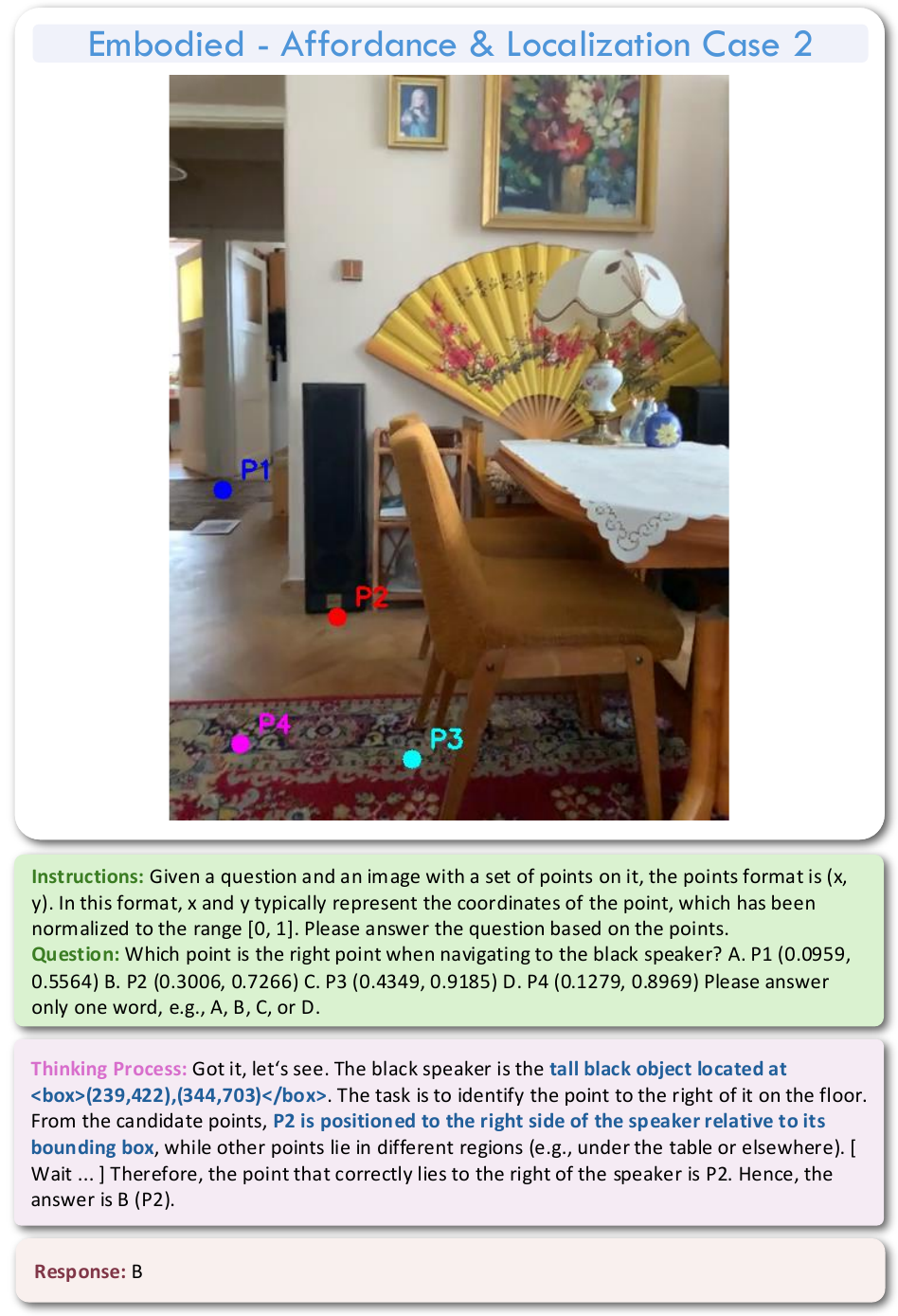}

    \vspace{0.5em}
    \small Fig.B11: \textbf{Example of navigation-oriented localization.} 
\end{center}

\clearpage
\noindent
\textbf{Task Planning Verification.}

\begin{center}
    \includegraphics[width=0.92\textwidth]{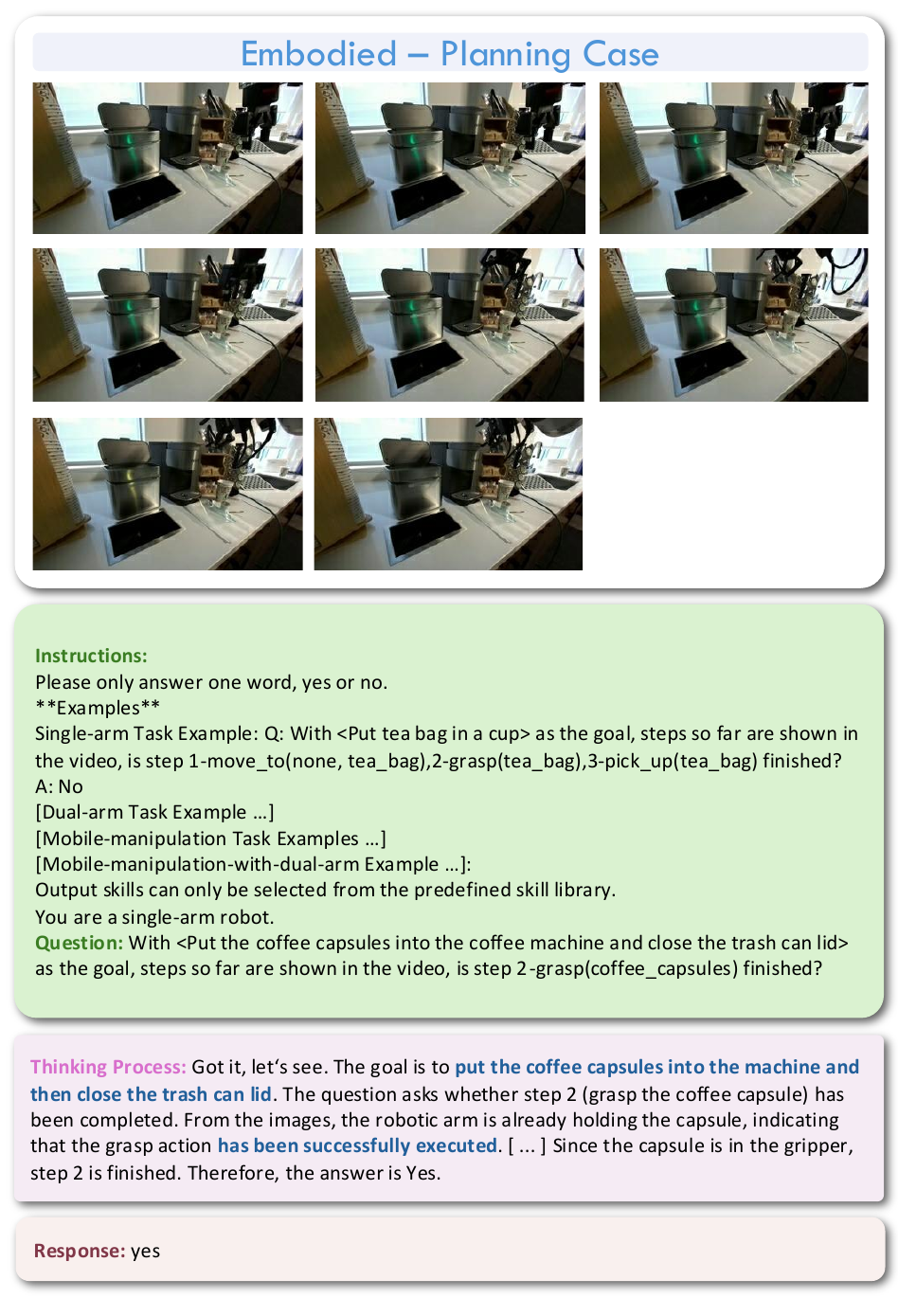}

    \vspace{0.5em}
    \small Fig. B12: \textbf{Example of task planning verification.} 
\end{center}

\clearpage
\noindent
\textbf{Embodied Question Answering.}

\begin{center}
    \includegraphics[width=0.92\textwidth]{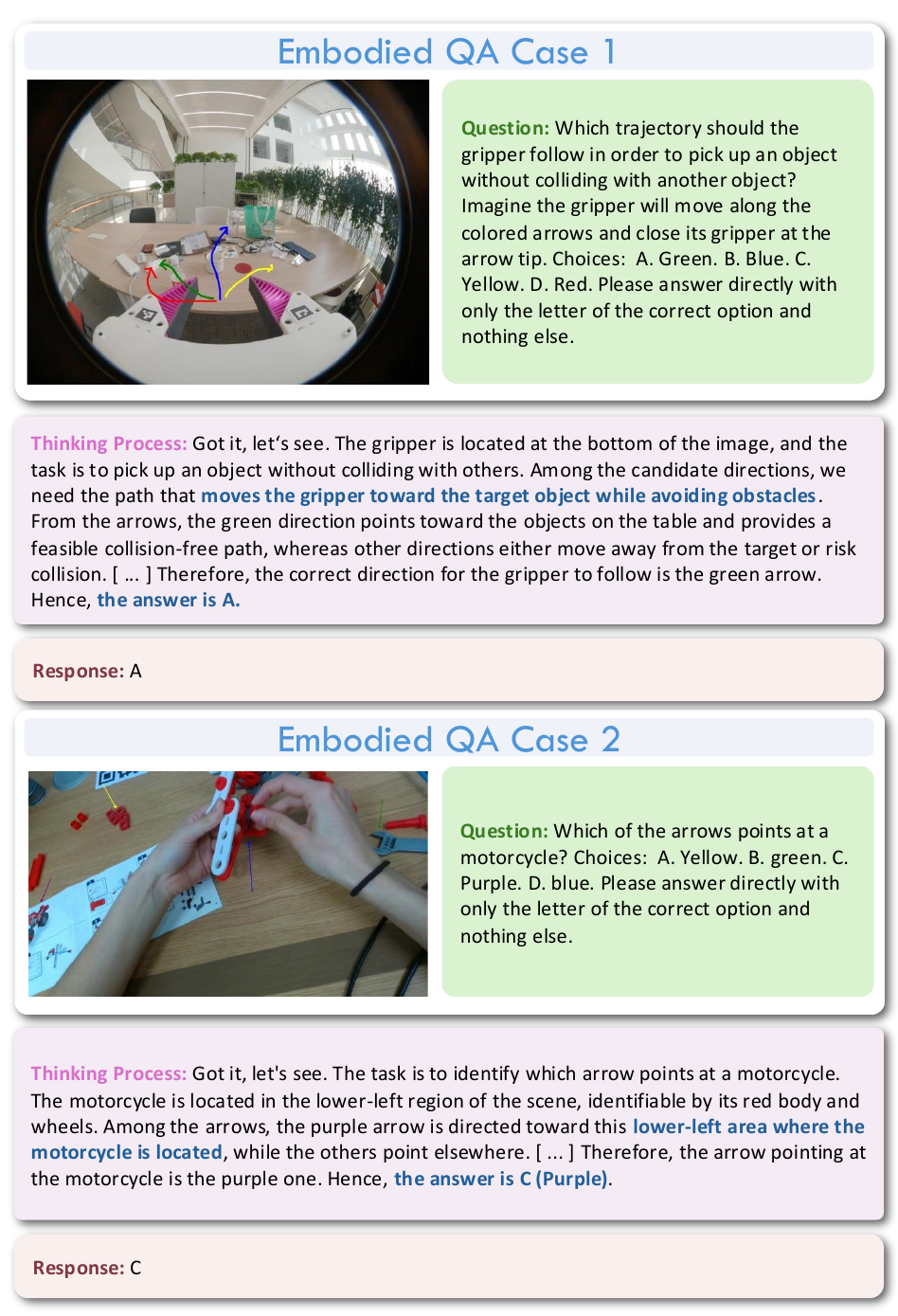}

    \vspace{0.5em}
    \small Fig.B13: \textbf{Examples of embodied question answering.}  
\end{center}

\clearpage
\noindent
\textbf{Sorting-based Reasoning Task.}

\begin{center}
    \includegraphics[width=0.92\textwidth]{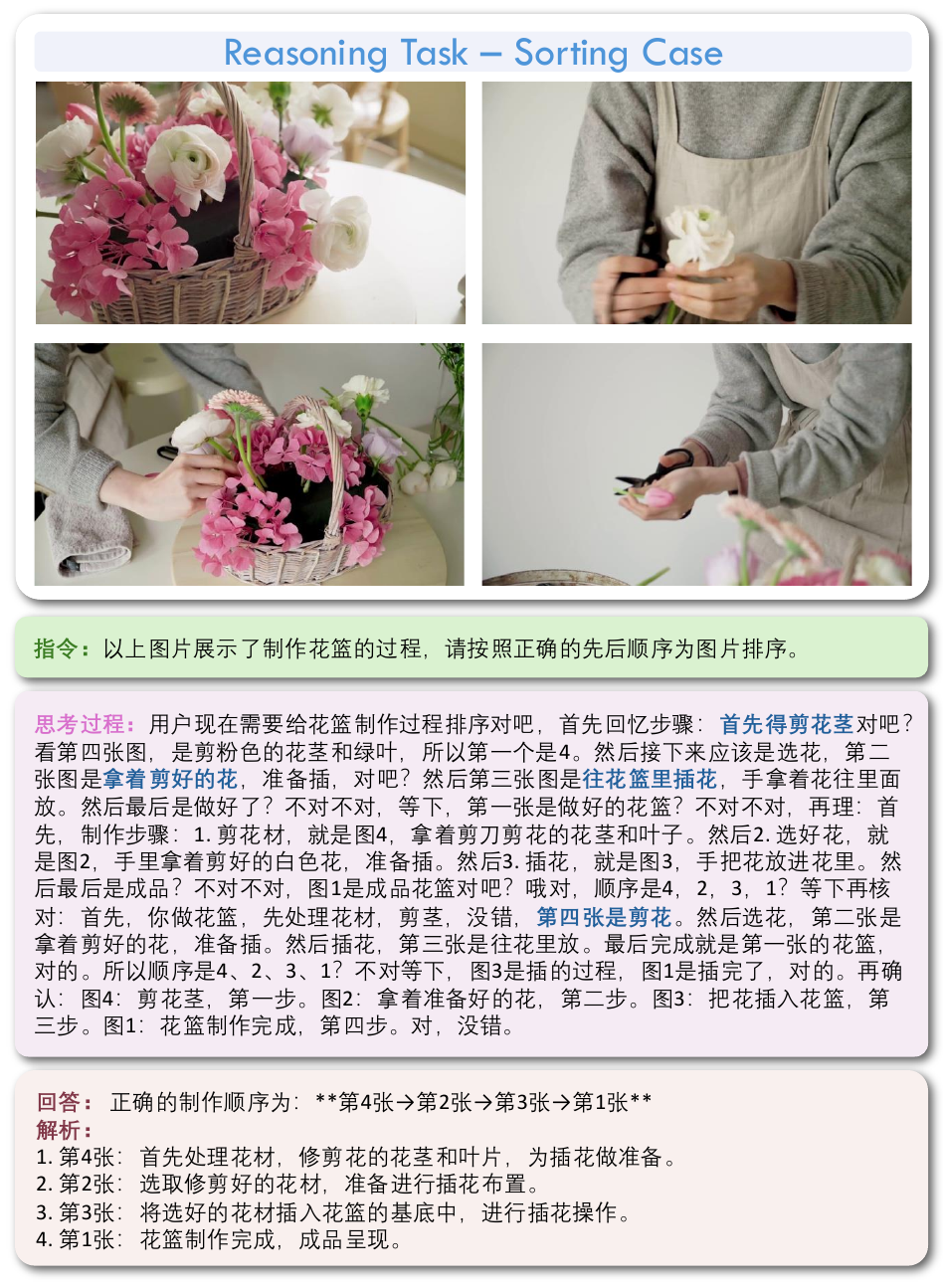}

    \vspace{0.5em}
    \small Fig.B14: \textbf{Example of sorting-based reasoning task.} 
\end{center}

\clearpage
\noindent
\textbf{Counting-based Reasoning Task.}

\begin{center}
    \includegraphics[width=0.92\textwidth]{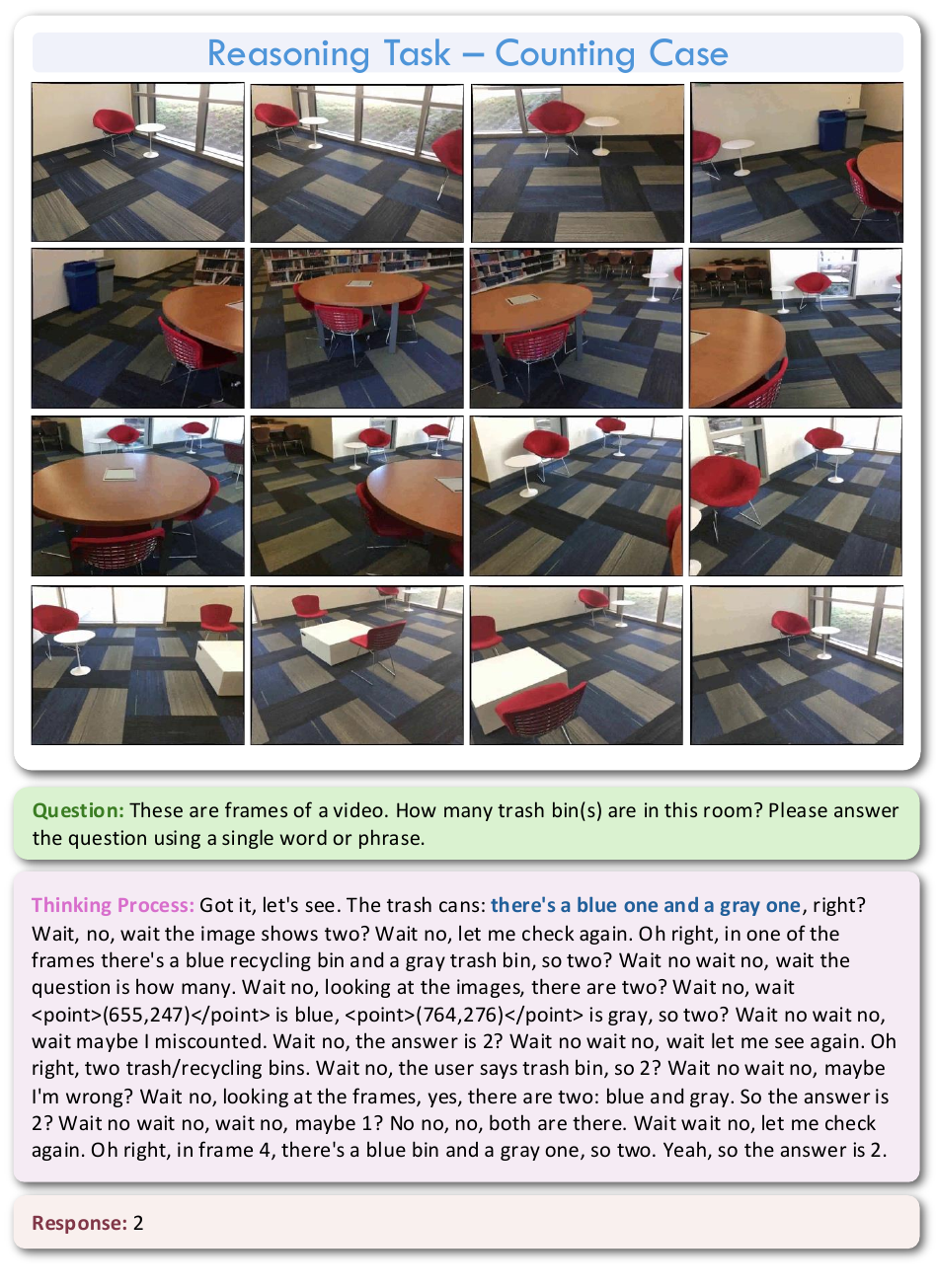}

    \vspace{0.5em}
    \small Fig.B15: \textbf{Example of counting-based reasoning task.} 
\end{center}

\clearpage
\noindent
\textbf{General Visual Question Answering Task.}

\begin{center}
    \includegraphics[width=0.92\textwidth]{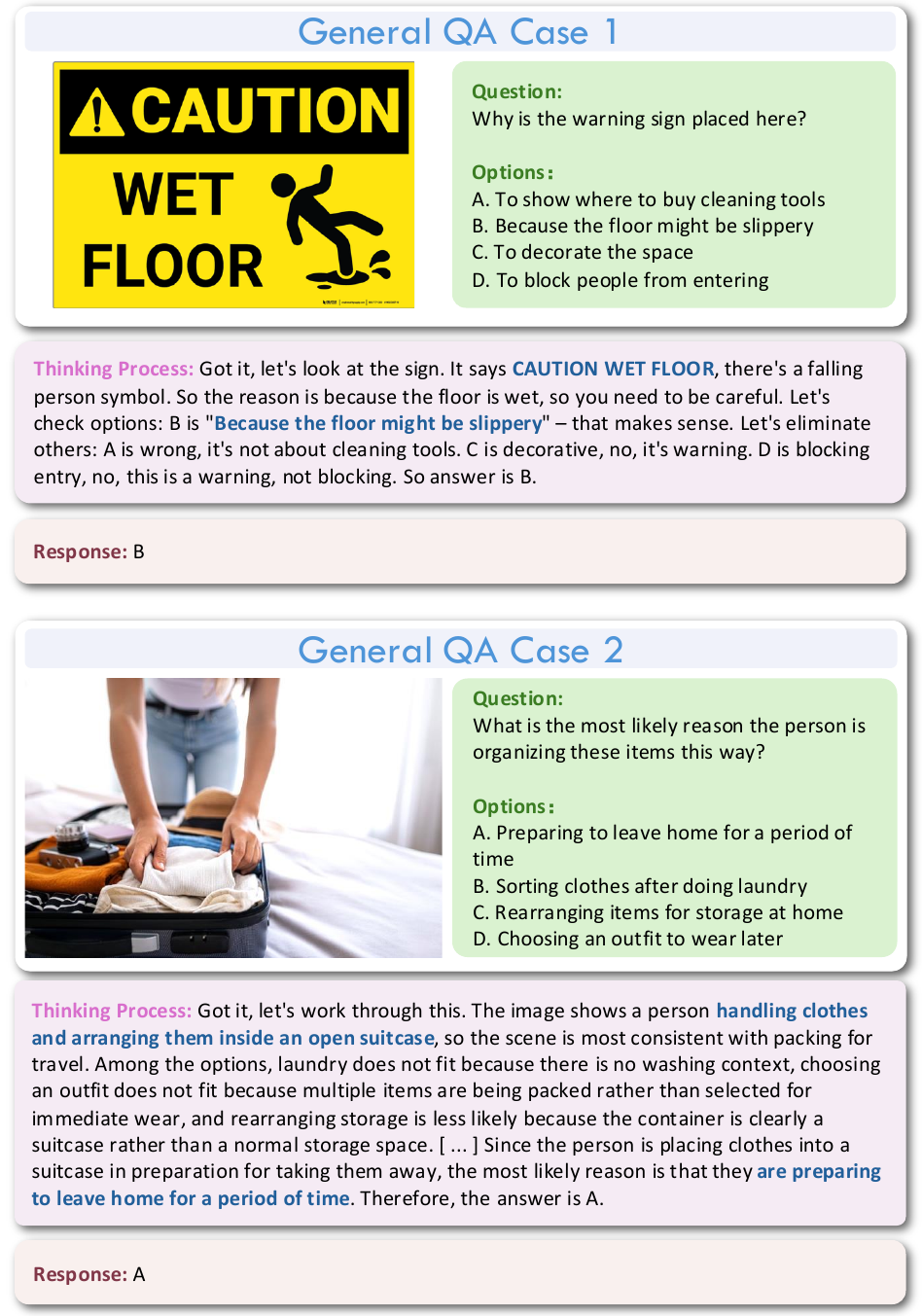}

    \vspace{0.5em}
    \small Fig. B16: \textbf{Examples of general visual question answering tasks.} 
\end{center}